\documentclass{article}

\usepackage[preprint,nonatbib]{neurips_2022}

\usepackage[utf8]{inputenc} 
\usepackage[T1]{fontenc}    
\usepackage{hyperref}       
\usepackage{url}            
\usepackage{booktabs}       
\usepackage{amsfonts}       
\usepackage{nicefrac}       
\usepackage{microtype}      
\usepackage{graphicx}
\usepackage{epstopdf}
\usepackage{wrapfig}
\usepackage{subcaption}
\usepackage{multirow}
\usepackage{tikz}
\usepackage{comment}
\usepackage{amsmath,amssymb} 
\usepackage{color}
\usepackage{hyperref} 
\usepackage[ruled,linesnumbered]{algorithm2e}

\title{Revisit Similarity of Neural Network Representations From Graph Perspective}

\author{%
  Zuohui Chen\thanks{Authors with $^{\ast}$ contributed equally to this work.} \\ 
  Institute of Cyberspace Security\\
  Zhejiang University of Technology\\
  Hangzhou, 310023, P.R. China\\
  \texttt{czuohui@gmail.com} \\
  \AND
  Yao Lu$^{\ast}$ \\ 
  Institute of Cyberspace Security\\
  Zhejiang University of Technology\\
  Hangzhou, 310023, P.R. China\\
  \texttt{yaolu.zjut@gmail.com} \\
  \And
  JinXuan Hu \\
  Institute of Cyberspace Security\\
  Zhejiang University of Technology\\
  Hangzhou, 310023, P.R. China\\
  \texttt{zjuthjx@163.com} \\
  \And
  Wen Yang \\
  Institute of Cyberspace Security\\
  Zhejiang University of Technology\\
  Hangzhou, 310023, P.R. China\\
  \texttt{wenyang.zjut@outlook.com} \\
  \And
  Qi Xuan\thanks{Qi Xuan is the corresponding author.}\\
  Zhejiang University of Technology\\
  Hangzhou, 310023, P.R. China\\
  \texttt{xuanqi@zjut.edu.cn} \\
  \And
  Zhen Wang\\
  School of Artificial Intelligence, Optics and Electronics (iOPEN)\\
  Northwestern Polytechnical University\\
  Xi’an 710072, P.R. China\\
  \texttt{zhenwang0@gmail.com} \\
  \And
  Xiaoniu Yang\\
  Institute of Cyberspace Security\\
  Zhejiang University of Technology\\
  Hangzhou, 310023, P.R. China\\
  \texttt{yxn2117@126.com} \\
}

\SetKw{Continue}{continue}

\begin{document}

\title{Revisit Similarity of Neural Network Representations From Graph Perspective}

\maketitle
\begin{abstract}
Understanding the black-box representations in Deep Neural Networks (DNN) is an essential problem in deep learning. In this work, we first revisit the state-of-the-art similarity index Centered Kernel Alignment (CKA) and find that it implicitly constructs graphs to obtain similarity. Nevertheless, the constructed graphs are not fully utilized since it does not realize their existence. To address this issue, we propose Graph-Based Similarity (GBS) to measure the similarity of DNN representations. GBS explicitly constructs graphs with the feature relationship between input samples and measures the correlation based on graphs. The graph not only provides a better similarity measurement but also contains rich information about the representation. Our experiments indicate that CKA also benefits from the sparse connection used in GBS, showing better performance in the sanity check. Compared with CKA, GBS provides a more fine-grained similarity reflection on the model structure and its motif can reflect the clustering degree of representations, which is also available for identifying the functional similarity of features. We also exploit GBS to reveal the changes in representations in adversarial attacks.
\footnote{We release our code at \url{https://github.com/implicitDeclaration/similarity}.}

\end{abstract}

\section{Introduction}
It is indisputable that Deep Neural Network (DNN) has achieved the most advanced performance in the fields like computer vision~\cite{ranftl2021vision}, natural language processing~\cite{brown2020language}, etc. In most machine learning tasks, a DNN learns feature representations end to end. It is still an open problem on how to understand the learned representations. A fundamental and initial step to this problem is measuring the similarity of different representations. Recent studies have revealed many interesting findings by similarity index~\cite{raghu2021vision,nguyen2020wide,nguyen2022origins}, between which the most recognized similarity index is Centred Kernel Alignment~\cite{kornblith2019similarity,cortes2012algorithms} (CKA). 

The success of CKA is attributed to the usage of relationships between samples~\cite{kornblith2019similarity}. Its power has also been validated in other deep learning domains. For example, knowledge distillation with samples' relationship is more robust to network changes~\cite{liu2019knowledge}, adversarial examples can be detected by their relationship with other clean examples~\cite{abusnaina2021adversarial}, and using samples relationship also provides improved performance on few-shot learning~\cite{sung2018learning}.
Hence we uncover the DNN representation similarity by the relationship of the samples. An intuitive way to mine the relationship between individuals is using the graph, which has been proved to be an effective tool in the deep learning community~\cite{you2020graph,zhou2021distilling}.


However, CKA does not realize the nature of this relationship constitutes the graph, nor uses any graph properties, while the graph contains rich information about the representation. To address the problem, we propose Graph-Based Similarity (GBS), which measures the representation similarity based on the graph. CKA can be regarded as a special case under our framework. Specifically, a batch of samples with a balanced distribution of labels are fed into the target model to obtain the features of each layer, then we construct a weighted undirected graph where each input is the node and the cosine similarity between each input pair is the edge weight. Finally, the similarity is calculated between graphs.

The constructed graphs not only can be used to measure the similarity but also contains rich information about the representation. In this work, we show that the ratio of triangle motifs, as a kind of recurring, significant patterns of interconnections in a graph, is closely related to the clustering degree of representations. This relationship can help us to identify whether the input features can achieve classification function i.e., functional similarity (\autoref{Sec:Related Works}).


Our contribution are summarized as follows:
\begin{enumerate}
\item We revisit CKA from a graph perspective and demonstrate that CKA uses the first-order neighbor information of its implicitly constructed graphs to obtain the similarity. Experiment shows that when using a sparser graph (CKA uses a fully connected graph), CKA achieves better performance in the sanity check.
\item To leverage the power of the graph and improve the similarity measurement, we propose Graph-Based Similarity (GBS). GBS is invariant to orthogonal transformation and isotropic scaling but not invertible linear transformation. It outperforms CKA at matching different model layers and provides a more fine-grained reflection of model structure. 
\item The explicitly constructed graph bridges the gap between representational similarity and functional similarity. We can use the similarity between graphs to measure the representational similarity while using motifs of the graph to measure the functional similarity. 
\end{enumerate}

\section{Related Works}
\label{Sec:Related Works}
Previous works proposed many indexes that reflect DNN similarity from different aspects. In this section, we give a short review of them and discuss what attributes a good similarity index should have. 
\subsection{Similarity of Representation}
The similarity of DNN features~\footnote{Both feature and representation refer to the output of hidden layers in DNNs.} can be generalized into representational similarity and functional similarity~\cite{csiszarik2021similarity}, which define similarity in representations and functions that process these representations, respectively. 

Formally, given two representations $\textbf{X}$ and $\textbf{Y}$, representational similarity $sim_r(\textbf{X}, \textbf{Y})\in [0, 1]$ is defined by calculating statistics on two different data embeddings. The functional similarity concerns if the representation can realize its function, which is usually achieved by model stitching~\cite{csiszarik2021similarity, bansal2021revisiting}. Concretely, two networks $U=u_L\circ \cdots \circ u_1$ and $W = w_L\circ \cdots \circ w_1$ are of the same structure but initialized with different random seeds, where $u_m: \textbf{F}_{m-1}\rightarrow \textbf{F}_m$ and $w_n : \textbf{F}_{n-1}\rightarrow \textbf{F}_n$ denote hidden layers of $U$ and $W$ respectively. Model stitching uses a stitch layer $S: \textbf{F}_{W,n}\rightarrow \textbf{F}_{U,m}$ to transform the output of $W$ at layer $n$ to be a suitable input for the corresponding layer $m$ in $U$ (since $U$ and $W$ have the same structure, $m=n$). The final stitched model is $S_{WU}=U_{L,m}\circ S\circ W_{n,1}$, where $U_{L,m}= u_L\circ \cdots  \circ u_m$ and $W_{n,1}=w_n \circ \cdots \circ w_1$. The stitching layer $S$ is optimized by minimizing the cross entropy between the output of $S_{WU}$ and ground truth label.

Previous works on representational similarity focus on comparing the DNN hidden layer's output, by decomposition of matrix~\cite{hardoon2004canonical,raghu2017svcca,morcos2018insights} or kernel function~\cite{kornblith2019similarity}. Kornblith et al.~\cite{kornblith2019similarity} analyzed various similarity indexes, including Canonical Correlation Analysis~\cite{hardoon2004canonical} (CCA), singular vector CCA~\cite{raghu2017svcca}, projection-weighted CCA~\cite{morcos2018insights}, and CKA~\cite{kornblith2019similarity,cortes2012algorithms}. They came to conclusion that CKA is the best representational similarity index among them through theoretical provement and experiments.

Recent literature discusses the similarity of representations from a variety of perspectives. Williams et al.~\cite{williams2021generalized} proposed representational dissimilarity to identify relationships between neural representations. Cui et al.~\cite{cui2022deconfounded} proposed that CKA can be improved with deconfounding, which will increase the resolution of detecting semantically similar neural networks. Bansal et al.~\cite{bansal2021revisiting} used model stitching to validate that self-supervised and supervised representation are functionally similar. Csisz{\'a}rik et al.~\cite{csiszarik2021similarity} proposed the concept of representational similarity and functional similarity. They showed that the representational similarity index is not necessarily indicative of functional similarity. 

Overall, CKA is generally recognized as the best representational similarity index and widely adopted in the downstream tasks~\cite{raghu2021vision,nguyen2020wide,nguyen2022origins}. Thus we focus on the comparison with CKA in this work. 

\subsection{How to identify a good similarity index?}
Kornblith et al.~\cite{kornblith2019similarity} proposed that similarity should be invariant to Orthogonal Transformation (OT) and Isotropic Scaling (IS) but not Invertible Linear Transformation (ILT).

Assuming $s(\cdot,\cdot)$ denotes the similarity index, $\textbf{X}$ and $\textbf{Y}$ are the representations. Being invariant to OT means $s(\textbf{X},\textbf{Y})=s(\textbf{X}\textbf{U},\textbf{Y}\textbf{V})$ for any full-rank orthogonal matrices $\textbf{U}$ and $\textbf{V}$. It implies that the index is invariant to permutation, which is necessary considering the symmetries of neural networks~\cite{orhan2017skip}. Being invariant to IS asks for $s(\textbf{X},\textbf{Y})=s(\alpha \textbf{X},\beta \textbf{Y})$ for any $\alpha,\beta \in \mathbb{R}^+$, i.e., the index is not affected by rescaling of individual features. Invariant to ILT means $s(\textbf{X},\textbf{Y}) = s(\textbf{X}\textbf{A},\textbf{Y}\textbf{B})$ for any full rank $\textbf{A}$ and $\textbf{B}$. As proved in \cite{kornblith2019similarity}, if a similarity index is invariant to ILT, it will give the same result for any representation of width greater than or equal to the dataset size. Such a result is not desired because this situation widely exists in convolutional networks~\cite{lee2017deep,springenberg2014striving}. 

In addition to the invariant properties, we also wish to validate the superior of an index by experiment. Kornblith et al. introduced a sanity check for similarity index in \cite{kornblith2019similarity}. Specifically, given a pair of architecturally identical models trained with the same dataset but different random initialization, the similarity index should assign the highest score for the architecturally corresponding layers. In addition to finding corresponding layers, a good similarity index should also reflect the model architecture. We can draw a confusion matrix to compare the similarity between each pair of layers in the model, where the building blocks of the model can be visualized.

\section{Method}
\label{Sec:Method}

In this section, we first revisit CKA from a graph perspective, then we propose GBS, a graph-based similarity index that unlocks the potential of graph.
\subsection{CKA from Graph Perspective}
As claimed by Kornblith et al~\cite{kornblith2019similarity}, CKA first measures the similarity between every pair of examples in their representation separately, and then compares the similarity structure. This process can be understood as the construction and comparison of graphs, which is intuitive with the linear kernel. Given two representations $\textbf{X}$ and $\textbf{Y}$, the linear kernel Hilbert-Schmidt Independence Criterion~\cite{gretton2005measuring} (HSIC) is
\begin{equation}
    HSIC(K_X, K_Y) = \frac{1}{(N-1)^2}tr(K_X\textbf{J}K_Y\textbf{J})=\frac{1}{(N-1)^2}tr(\textbf{X}\textbf{X}^T\textbf{J} \textbf{Y}\textbf{Y}^T\textbf{J}),
\end{equation}

where $N$ is the number of examples, $J=I_N-\frac{1}{N}\textbf{1}$ is the centering matrix, and $K_X(\textbf{X},\textbf{X})$ is the kernel. Representations $\textbf{X}\in \mathbb{R}^{N\times D}$ and $\textbf{Y}\in \mathbb{R}^{N\times D}$ are both obtained by $N$ examples, where $D$ is feature dimensions. 

Considering the physical meaning of $\textbf{X}$ and $\textbf{Y}$ that each row is a layer's output for different examples, linear HSIC embeds representations $\textbf{X}$ and $\textbf{Y}$ into two adjacent matrices $\textbf{A}_X=\textbf{X}\textbf{X}^T$ and $\textbf{A}_Y=\textbf{Y}\textbf{Y}^T$, where nodes are input examples and edges are the inner product of the examples outputs in the same layer. $\textbf{A}_X$ and $\textbf{A}_Y$ describe the relationships between input examples and the role of the centering matrix can be regarded as subtracting a constant from each item of the adjacent matrices, thus does not change the essence of its physical meaning. Assuming centered adjacent matrices are $\tilde{\textbf{A}}_X$ and $\tilde{\textbf{A}}_Y$, that both of them are weighted undirected graph, i.e., $\tilde{\textbf{A}}_X$ and $\tilde{\textbf{A}}_Y$ are symmetric. The diagonal of $\tilde{\textbf{A}}_X\tilde{\textbf{A}}_Y$ represents the inner product of first-order neighbor information between each node. HSIC then averages the elements in the diagonal as the final result. CKA further normalizes HSIC by 
\begin{equation}
    CKA(\textbf{X}, \textbf{Y})= \frac{HSIC(K_X, K_Y)}{\sqrt{HSIC(K_X, K_X)HSIC(K_Y, K_Y)}},
\end{equation}
because HSIC is a test statistic for independence instead of mutual information and normalization can induce invariance to isotropic scaling. 

Nevertheless, we argue that there is still room for improvement in graph construction and similarity comparison. The topology of the kernel-constructed graph is fully connected, which may bring noise for similarity measurement. Hence, pruning irrelevant edges may improve the performance. The linear HSIC with sparse connections is
\begin{equation}
   \frac{1}{(m-1)^2}tr(M(\textbf{A}_X, m)\textbf{J}_m M(\textbf{A}_Y, m)\textbf{J}_m),  
\end{equation}
where $m$ is the number of reserved edges, $M(\cdot, \cdot)$ represents reserving top $m$ edges in each row of the adjacent matrix, and $\textbf{J}_m=I_N-\frac{1}{m}\textbf{1}$. Meanwhile, considering that rows and columns of representations have clear physical meaning, the properties of the graph can reveal more information to us.
\begin{figure*}[t!]
\centering
\includegraphics[width=0.90\textwidth]{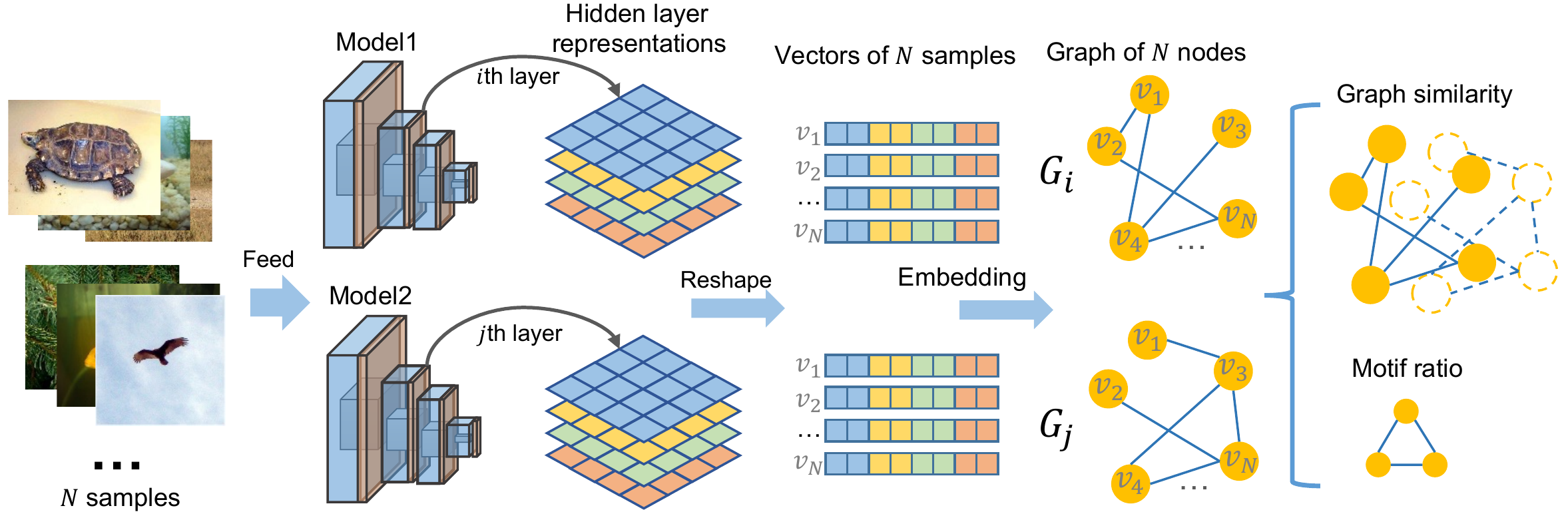}
\caption{Pipeline of computing GBS.}
\label{Fig:method}
\end{figure*}

\subsection{Graph-Based Similarity}
As shown in \autoref{Fig:method}, GBS requires a batch of samples to obtain the hidden layer representations. After that, representations of a layer are embedded in a graph, then we compare the similarity between graphs. 

Given a set of samples $\mathcal{X} = \{ \mathit{x}_1, \mathit{x}_2, \dots , \mathit{x}_N \}$ where $N$ represents the number of samples derived from the dataset, we feed them into the target model with fixed parameters to obtain neural network hidden representations $\textbf{F}_i \in \mathbb{R}^{N \times \mathit{C}_i \times \mathit{W}_i \times \mathit{H}_i}$, $ i = 1, \cdots, L $, where $L$ is the number of layers, $\mathit{W}_i$ and $\mathit{H}_i$ are width and height of feature maps, and $\mathit{C}_i$ is the number of channels in $i$th layer, respectively. Note that we treat a building block containing multiple convolutional layers as a layer for ResNet~\cite{he2016deep} and treat a single convolutional layer as a layer for VGGNet~\cite{simonyan2014very}. Then we reshape the obtained representation to reduce its dimension, i.e., $\mathbb{R}^{N \times \mathit{C}_i \times \mathit{W}_i \times \mathit{H}_i} \rightarrow \mathbb{R}^{N \times \mathit{M}_i}$, where $\mathit{M}_i = \mathit{C}_i \times \mathit{W}_i \times \mathit{H}_i$. 

We define the graph of $i$th layer corresponding to the input $\mathcal{X}$ as $\mathcal{G}_i=(\mathcal{V}, \mathcal{E})$ by a node set $\mathcal{V}=\{v_1, ..., v_N\}$ and an edge set $\mathcal{E}\subseteq \{(v_n, v_m)|v_n, v_m\in \mathcal{V}\}$. The number of nodes in $\mathcal{G}_i$ is determined by the number of samples $N$, which is independent of the feature map size. Each node $v_k$ corresponds to its input $x_k$ and the feature of $v_k$ is $\textbf{F}_{i|k}$. The edge weight between $v_n$ and $v_m$ is calculated by the cosine similarity of their features
\begin{equation}
\mathit{e}_{nm} = \dfrac{\textbf{F}_{i|n} \cdot \textbf{F}^T_{i|m}}{\Vert \textbf{F}_{i|n} \Vert _2 \cdot \Vert \textbf{F}_{i|m} \Vert _2}.
\label{eq:cos}
\end{equation}
Compared with linear CKA, we normalized the edge weight, only considering the angle between representations. Because the amplitude of DNNs representations does not determine their performance, e.g., a binary neural network~\cite{qin2020binary} can also obtain high accuracy on various tasks. 

Then we obtain the similarity of layer $i$ and layer $j$ by computing the similarity $s_{ij}$ between $\mathcal{G}_i$ and $\mathcal{G}_j$. 
We use Layer Similarity~\cite{zhang2020measuring} (LSim) to calculate the similarity between layers. LSim is proposed to measure the similarity between layers of complex and multiplex networks. Assuming the adjacent matrices of $\mathcal{G}_i$ and $\mathcal{G}_j$ are $\textbf{A}_i$ and $\textbf{A}_j$, the similarity is calculated by 
\begin{equation}
s_{ij} = \frac{1}{N}\sum_{k=1}^N f_{cos}(\textbf{a}_{i,k}, \textbf{a}_{j,k}),
\label{eq:gsimilarity}
\end{equation}
where $\textbf{a}_{i,k}$ is the row vector of $\textbf{A}_i$ and $f_{cos}(\cdot)$ is the cosine similarity between two vectors.

However, when $N$ is large, a fully connected graph will induce noise in the result, since many irrelevant nodes are also connected. To obtain a better result, we only connect the top $k$ nodes ($k$ equals the degree of the node) with the highest similarity at the step of graph construction.

As introduced in \autoref{Sec:Related Works}, a good similarity index should be invariant to orthogonal transformation and isotropic scaling but not invertible linear transform. GBS satisfies the requirements and we provide the proof in the appendix. 
\subsection{Motif and Feature Clustering}

\begin{figure}
    \centering
    \begin{subfigure}[b]{0.47\textwidth}
    \includegraphics[width=0.9\textwidth]{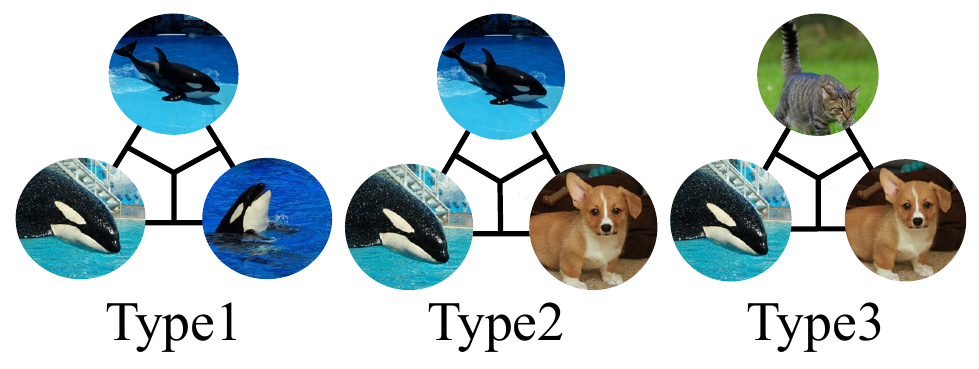}
    \caption{Motif types.}
    \label{Fig:motif_type}
    \end{subfigure}
    \begin{subfigure}[b]{0.47\textwidth}
    \includegraphics[width=0.9\textwidth]{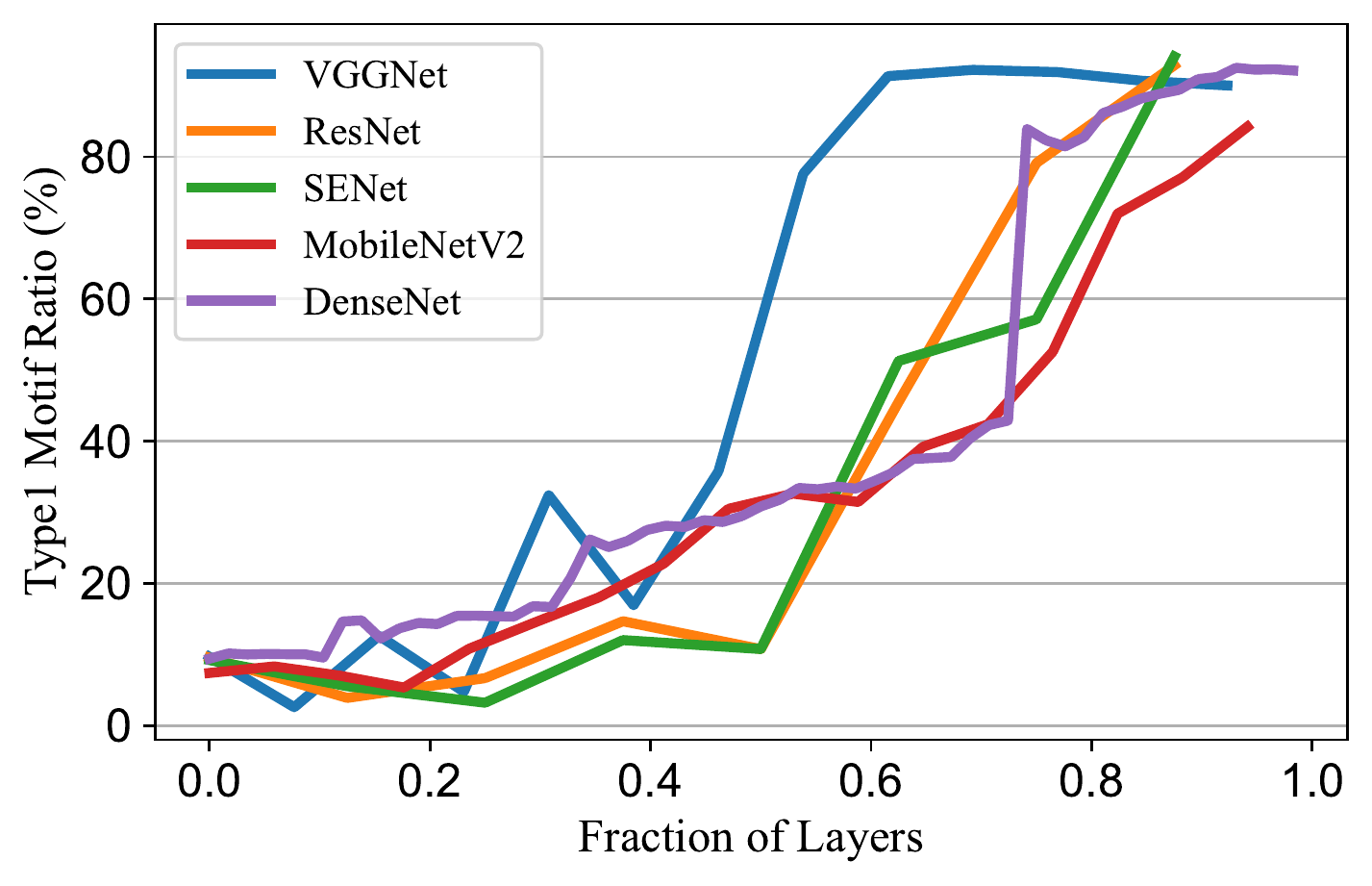}
    \caption{Variation of type1 motif ratio with layer depth.}
    \label{Fig:motif_ratio}
    \end{subfigure}
    
    \caption{Motifs in neural networks.}
    \label{Fig:motif_in_model}
\end{figure}

\begin{algorithm}[h]
    \caption{Breadth-First Motif Search}
    \label{Alg:bfs}
    \SetKwInOut{Input}{input}\SetKwInOut{Output}{output}
    
        \Input{A graph $\mathcal{G}=(\mathcal{V}, \mathcal{E})$ and graph size $N$ (number of nodes).}
        \Output{Number of different types of motifs.} 
        
        \For{$i$=$1$ \KwTo $N$}{
        Initialize a motif queue $\mathcal{Q}$ with an incomplete motif $q_0=\{v_i, \ , \ \}$\;
            \While{$\mathcal{Q}$ is not empty}{
                $q = \textit{dequeue}(\mathcal{Q})$\;
                \If{$q$ has 3 neighbors}{
                    Add one to the number of motifs of the corresponding type\;
                    \Continue
                }
                get $v_{current}$, which is the last node in $q$\tcp*{e.g., when $q=\{v_1, , \ \}$, $v_{current}=v_1$} 
                \For{$j$=$current$ \KwTo $N$}{
                    \If{$v_j$ has been visited}{
                     \Continue
                    } 
                    \If{$v_{current}$ is connected to $v_j$}{
                        $q_{new}=\textit{copy}(q)$\;
                        add $v_j$ to $q_{new}$\;
                        \textit{enqueue}($q_{new}$, $\mathcal{Q}$)\;
                    }
                }
            }
        }
\end{algorithm}

In addition to similarity measurement, we use the triangle motifs~\cite{watts1998collective} as the functional similarity index. Triangle motif consists of three nodes that are all fully connected. 


As shown in \autoref{Fig:motif_in_model} (a), there are three types of motifs according to their ground truth labels. Since previous works~\cite{caron2018deep,watanabe2019understanding,filan2021clusterability} indicate that features of the same category tend to cluster in deep layers, we focus on the ratio of the first type (type1) across all the motifs in a graph, i.e., samples with the same labels are connected.

As shown in \autoref{Alg:bfs}, we design a Breadth-First Search (BFS) based strategy to count the number of different types of triangle motifs. Specifically, we will traverse all nodes in the graph as the root node of searching. For each root node, we initialize a queue $\mathcal{Q}$ with an element $q_0=\{v_i, \ , \ \}$ that only has a root node. 
When $\mathcal{Q}$ is not empty, we do the following loop: \textbf{Step 1.} We dequeue $\mathcal{Q}$ to obtain $q$ and count the number of nodes in $q$. If the number of nodes is equal to 3, we first verify whether these nodes form a triangle motif. If they do, we classify the triangle motif according to the label of each node and execute the next loop, otherwise, execute step 2. \textbf{Step 2.} For the last node $v_{current}$ in $q$, we traverse all nodes whose node number is greater than $v_{current}$. If the node is connected with $v_{current}$ and has not been visited, we add it to $q$ and enqueue $q$ to $\mathcal{Q}$. According to the above steps, we can count the number of three types of triangle motifs and calculate the proportion of each type of triangle motif.

\autoref{Fig:motif_in_model} (b) shows the type1 motif ratio as a function of model depth in different models. It demonstrates that the ratio of type1 motif increases with the depth. Hence, we argue that the ratio of the type1 motif is an indicator of whether useful features have been extracted. 

\section{Experiments}
\label{Sec:Experiments}
\subsection{Experimental Settings}
We conduct our experiments on CIFAR10, CIFAR100~\cite{krizhevsky2009learning}, and ImageNet~\cite{5206848}. The three datasets contain 10, 100, and 1000 categories of images, respectively. The images size of CIFAR10 and CIFAR100 is $32\times 32$ while that of ImageNet is $224\times 224$. The networks we used are 10layer-CNN~\cite{springenberg2014striving}, VGGNet~\cite{simonyan2014very}, and ResNet~\cite{he2016deep}. Their detailed architectures are provided in the appendix.

\subsection{Ablation Analysis and Sanity Check}

Our ablation analysis focus on the influence of 1) the number of samples (graph size) and 2) the number of connected nodes (degree of the graph). We use the sanity check to evaluate the performance of GBS as described in \autoref{Sec:Related Works}. We use ten ResNet18, VGGNet16, and 10layer-CNN of different initialization trained on CIFAR10, and compute the average accuracy over all the layers. The initialization utilizes Kaiming normalization~\cite{he2015delving} with ten different random seeds. The average accuracy of three groups of models is above 89\%.

\begin{figure}[]
    \centering
    \includegraphics[width=0.9\textwidth]{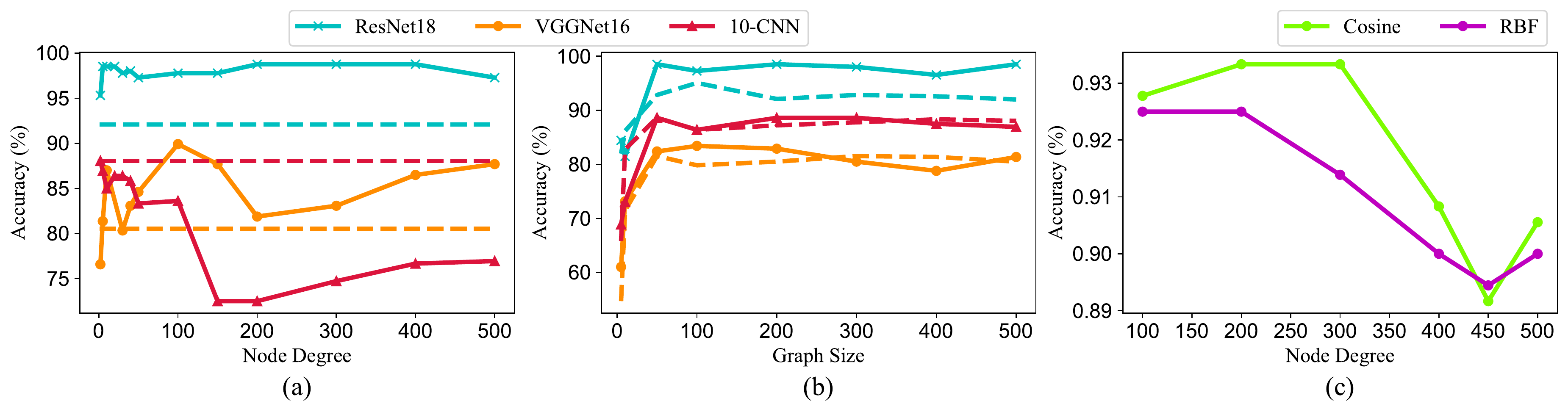}
    \caption{Sanity check accuracy using 10layer-CNN, VGGNet16, and ResNet18 on CIFAR10. In (a) and (b), the solid lines are the accuracy of GBS, while the dash lines are the accuracy of CKA; (c) shows the impact of kernels and sparsity, performed on 10layer-CNN.}
    \label{Fig:sanity}
\end{figure}

Given the intuition that an adequate number of nodes are beneficial to obtain a good similarity, we start with analyzing the influence of degree (top k connected nodes) under a fixed number of nodes (500). The 500 samples are randomly selected from the dataset that has a balanced label distribution. The result is shown in \autoref{Fig:sanity} (a), where the degree varies from 1 to 500 and we also calculated the unmodified linear CKA on 3 models. It shows that the degree of the generated graph has little impact on finding the layer correspondence. Even when the degree is set to 1, the accuracy only drops less than 6\% compared with the degree set to 10. When the degree increase to 500, i.e., using a fully connected graph, the accuracy slightly drops, which indicates that a fully connected graph may bring redundant information and is detrimental to the extraction of relevant information. Overall, GBS has better performance in the sanity check. Considering the computational cost, we set the degree to 5, i.e., only connect the top 5 most correlated nodes in the graph. 


\begin{figure*}[]
    \centering
    \begin{subfigure}[b]{0.47\textwidth}
           \centering
           \includegraphics[width=\textwidth]{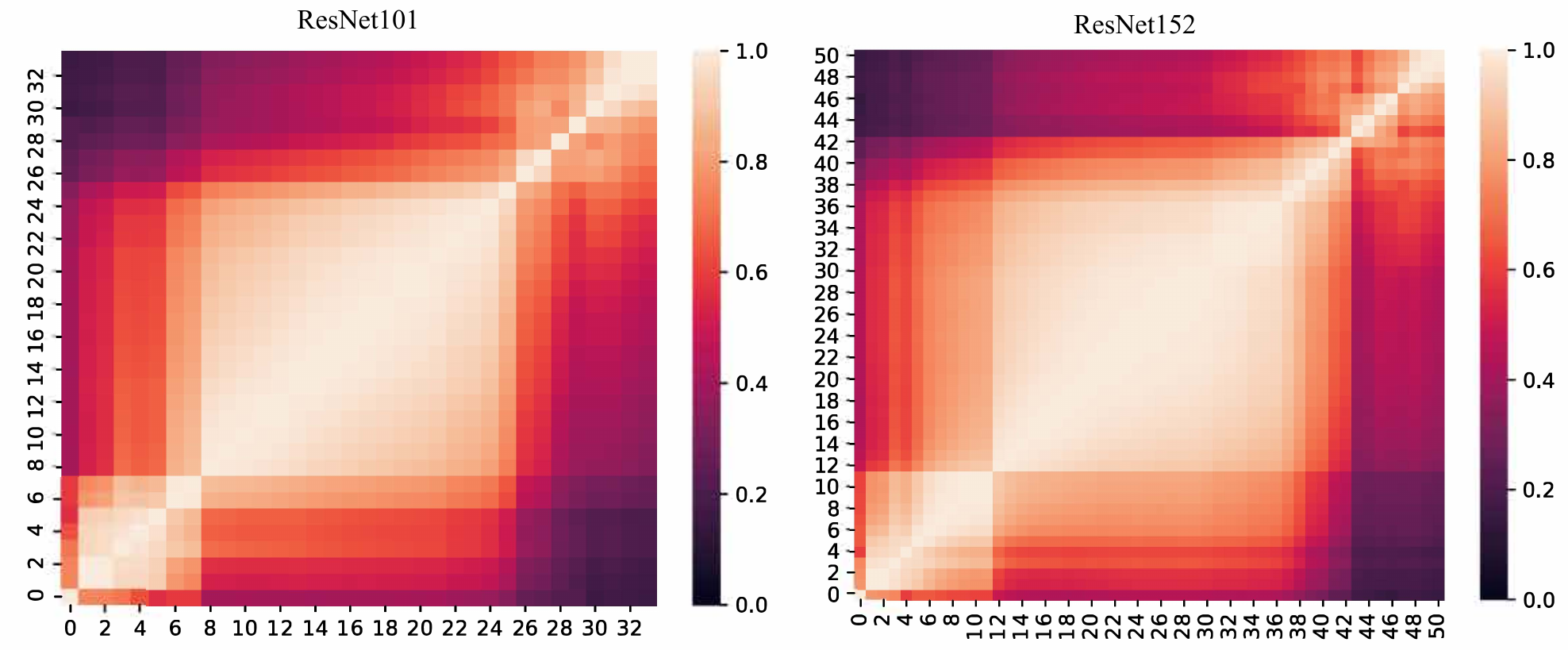}
           \includegraphics[width=\textwidth]{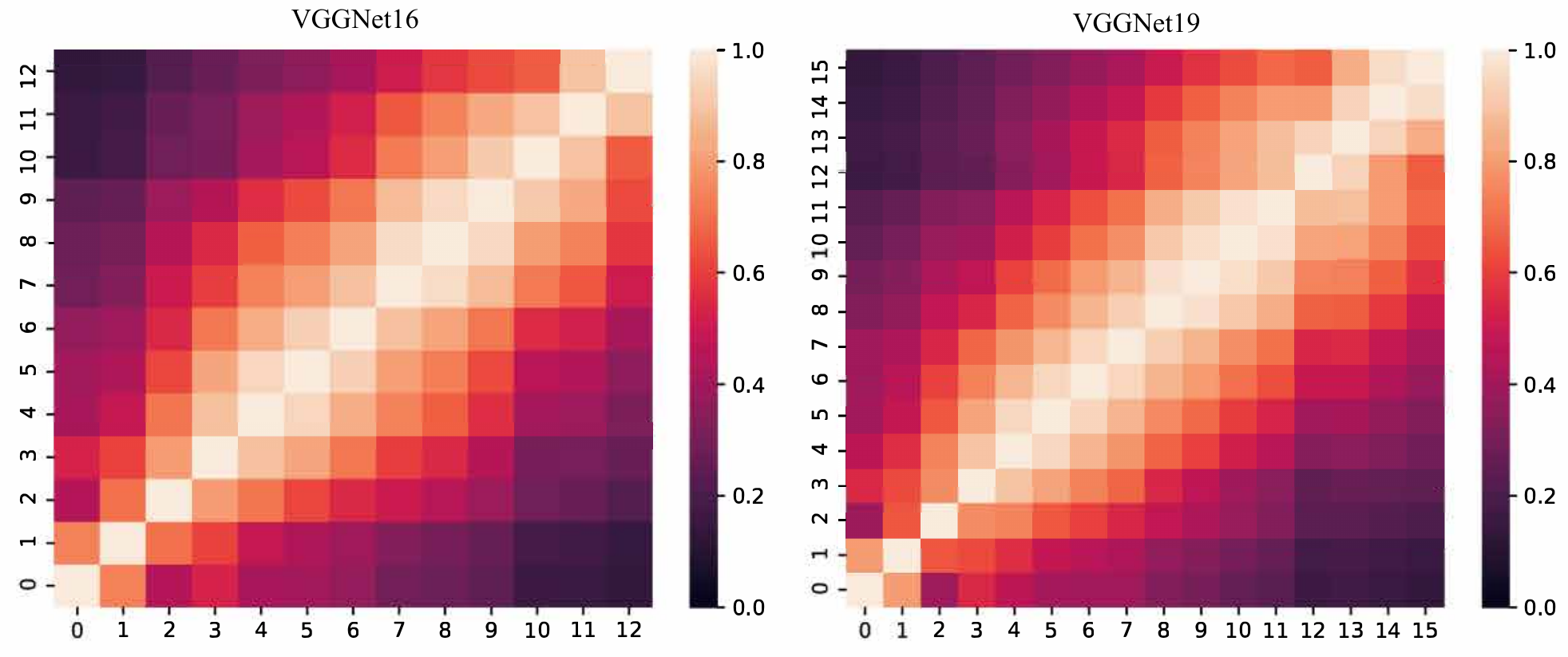}
           \caption{CKA}
    \end{subfigure}
    \begin{subfigure}[b]{0.47\textwidth}
            \centering
            \includegraphics[width=\textwidth]{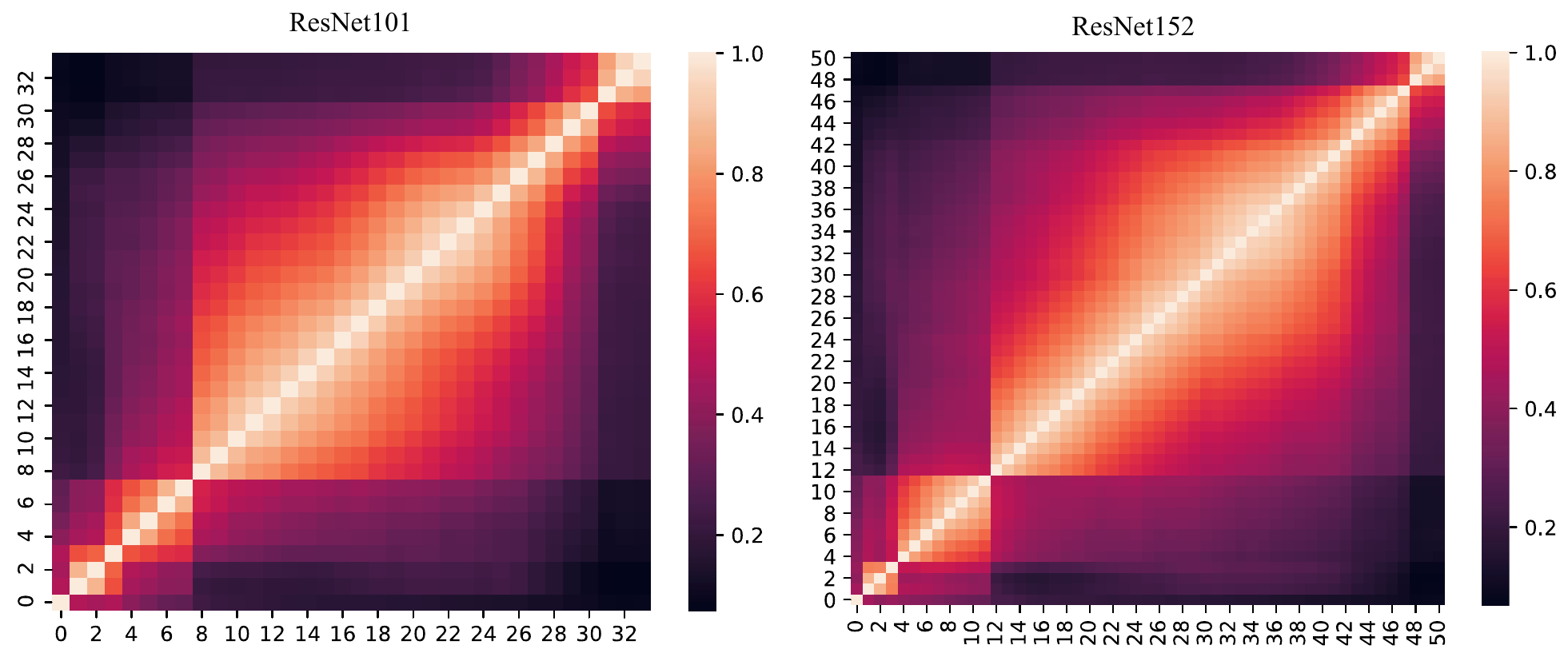}
            \includegraphics[width=\textwidth]{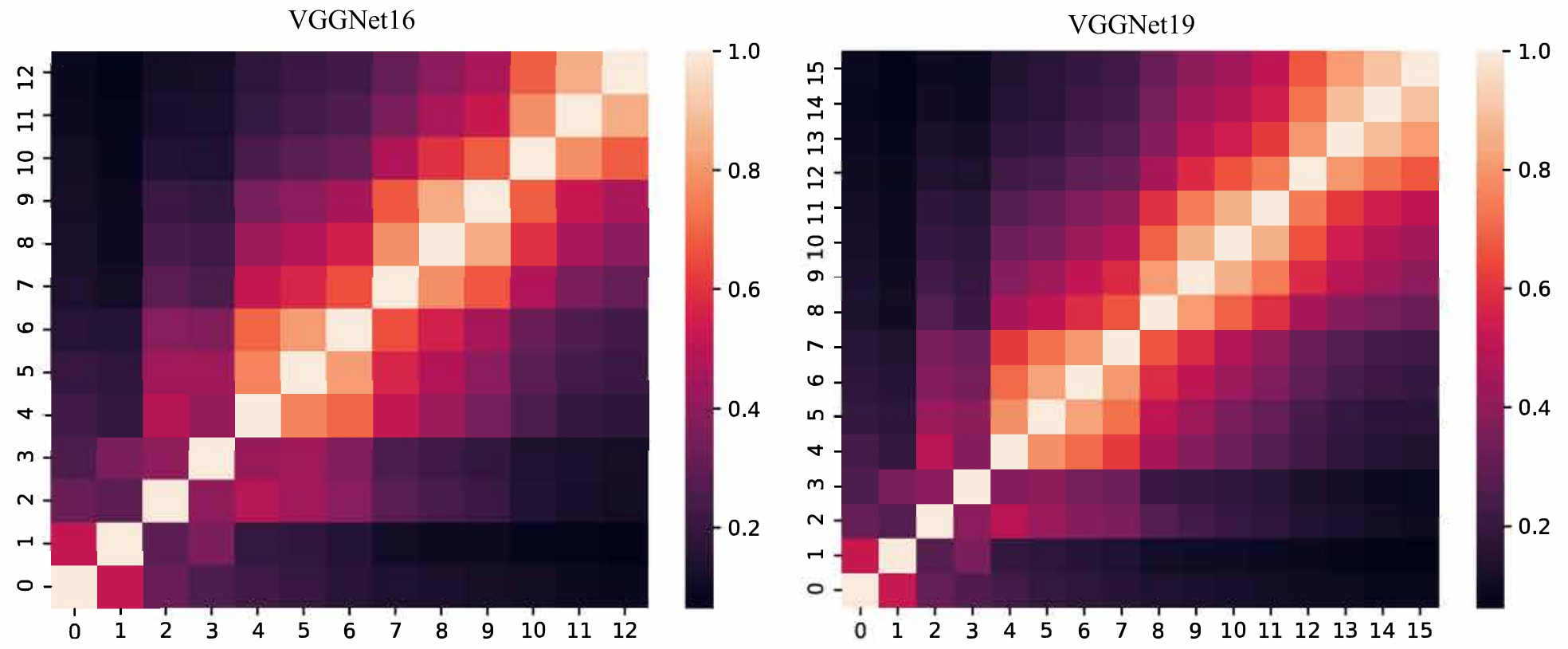}
            \caption{GBS}
    \end{subfigure}
    \caption{The similarity between the layers of VGGNets and ResNets on CIFAR10.}
    \label{Fig:pathology}
\end{figure*}
Then we evaluate the performance of GBS with the graph size (number of nodes in the graph) varies from 5 to 500, as shown in \autoref{Fig:sanity} (b). Note that the number of categories of CIFAR10 is 10. When the graph size is set to 5, i.e., less than the number of categories of CIFAR10, GBS still achieves accuracy over 61.0\% and 84.4\% on VGGNet16 and ResNet18, respectively. When the graph size is greater than 50, the accuracy hardly changes with it. 

We further show that the performance of CKA on sanity check can be improved by inducing sparsity and normalized edge weight in \autoref{Fig:sanity} (c). We use 500 input examples and 10layer-CNN. The Node Degree axis represents the reserved top $k$ edges during CKA embeds representations into graphs, where $500$ represents the unmodified CKA. It indicates that for RBF kernel and cosine kernel, sparsity can significantly improve the performance of CKA.

As shown in \autoref{Fig:pathology}, we use the confusion matrix to demonstrate the similarity of different layers within a model. The models in \autoref{Fig:pathology} are trained on CIFAR10, we also provide the results of CIFAR100 and ImageNet, together with the detailed model structures in the appendix. For both VGGNets and ResNets, grids in the confusion matrices exactly correspond to the size and depth of the block in the model, i.e., the similarity of layers inside the same block is higher than that of layers in different blocks. GBS has a more clear and more fine-grained presentation for the models' architecture compared with CKA. Within a grid, the color gradually darkens with the diagonal as the center, which means inside the same block, the greater the distance between layers, the lower the similarity. Because within a block, DNN extracts features layer by layer, the similarity gradually decreases accordingly. The residual connections deliver the features of different depths between the entry of blocks, making the representations of the current block significantly different from the post block.

In summary, the ablation analysis indicates that GBS is robust to the choice of graph size and degree. The sanity check implies that GBS is a more fine-grained similarity index than CKA. Increasing degree values slightly helps improve the performance of GBS. Considering the computational cost, an excessively large number of nodes and large degree values are unnecessary.

\subsection{Functional Similarity and Motifs}
We show that motif ratio can measure the functional similarity by two experiments, 1) stitching trained and randomly initialized models; and 2) stitching models trained by different amounts of data and epochs. ResNet18 and CIFAR10 are adopted in this experiment. All stitching layers are trained with 100 epochs, using SGD optimizer, and the learning rate is set to $0.1$.

\begin{figure*}[htbp]
	\centering
	\includegraphics[width=0.95\textwidth]{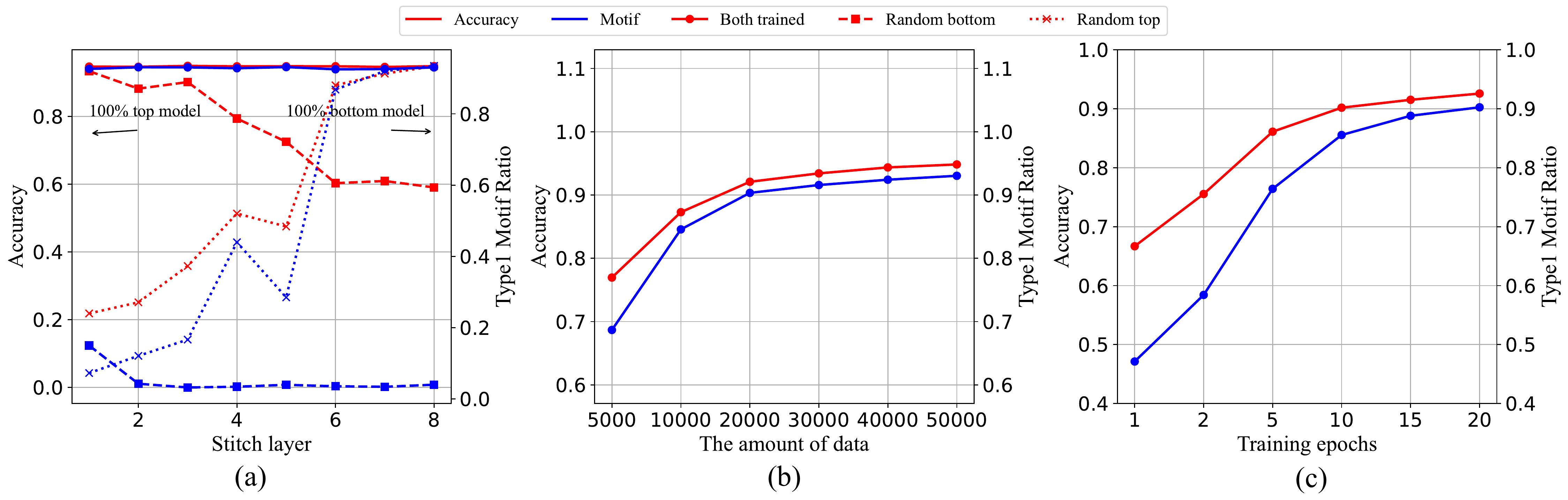}
	 \caption{Variation of motif with the (a) stitch layer, (b) training data, and (c) training epochs.}
	 \label{Fig:stitch_motif}
\end{figure*}

As shown in \autoref{Fig:stitch_motif}, we use color to distinguish accuracy and motif ratio and use different line styles to distinguish stitch cases. Red lines denote the accuracy after stitching and blue lines denote the motif ratio after stitching. Solid lines with circle denote stitching two models of different initialization, dash lines with rectangle denote stitching a trained model with a random bottom network, and dot lines with cross denote stitching a random top model with a trained model.

The functional similarity depends not only on whether the extracted features of the bottom model are effective but also on the strength of the top model that processes the features. In \autoref{Fig:stitch_motif} (a), when the top and bottom model matches (both trained), the motif ratio is high as they are highly functionally similar. For a random bottom model, since it can not provide useful representation, the motif, as well as functional similarity, is low at all the layers. A trained bottom model can provide informative representation for classification, nevertheless, the top model is a random model with a poor ability to process these representations, thus the motif, i.e., functional similarity increases with the layer depth.

Previous works~\cite{bansal2021revisiting} have shown that training with more data and epochs provides functionally more similar representations. We stitch a model trained with different samples and epochs to another fully trained model, to show how the motif varies with the amount of data and training epochs. As shown in \autoref{Fig:stitch_motif} (b) and \autoref{Fig:stitch_motif} (c), two models are stitched in the second block. The motif ratio is positively correlated with functional similarity, which indicates that the motif ratio is closely related to the functional similarity. Features with weak functionality decrease connections between same category samples, while functionally stronger features bring more dense connections.

Overall, our explicitly constructed graph bridges the gap between functional similarity and feature similarity. The similarity between graphs manifests the feature similarity while the graph motif is correlated with the functional similarity.

\subsection{Representation of Adversarial Samples }
\begin{figure*}[!t]
    \centering
    \includegraphics[width=0.95\textwidth]{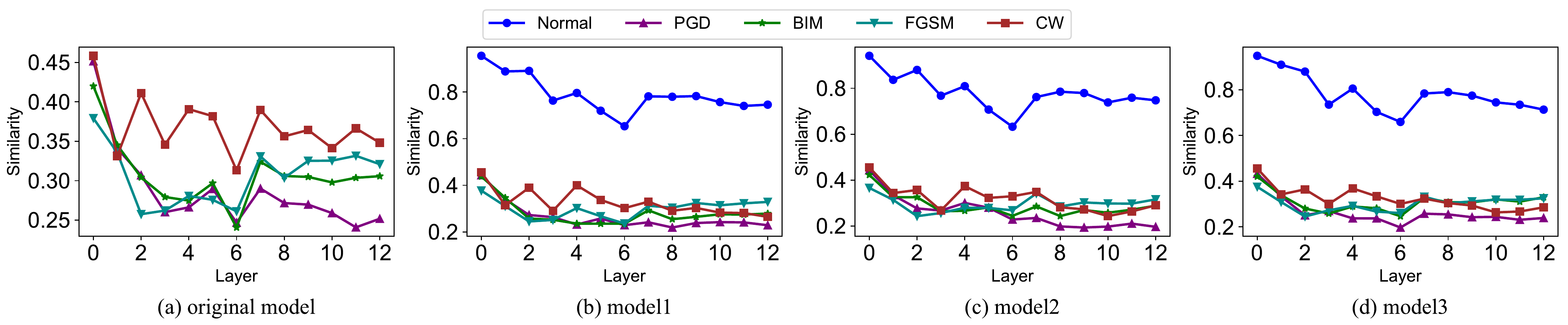}
    \caption{The representational similarity of adversarial samples transferring to different models. In (a), the similarity is calculated between normal and adversarial samples. In (b), (c), and (d), we also calculate the similarity between original model and other models using normal samples.}
\label{Fig:trans}
\end{figure*}
\begin{figure*}[t!]

    \centering
    \begin{subfigure}[b]{0.95\textwidth}
           \centering
           \includegraphics[width=\textwidth]{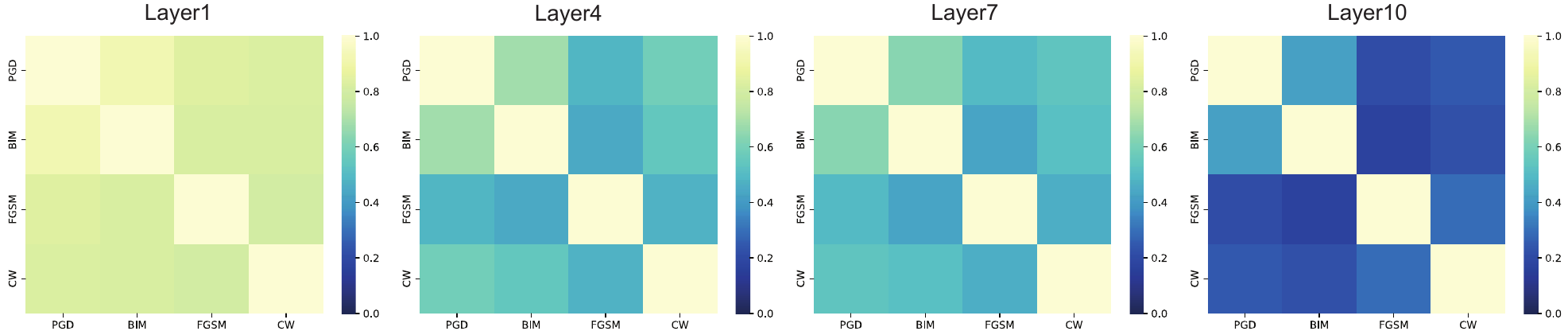}
           \caption{Targeted Attacks.}
    \end{subfigure}
    \begin{subfigure}[b]{0.95\textwidth}
            \centering
            \includegraphics[width=\textwidth]{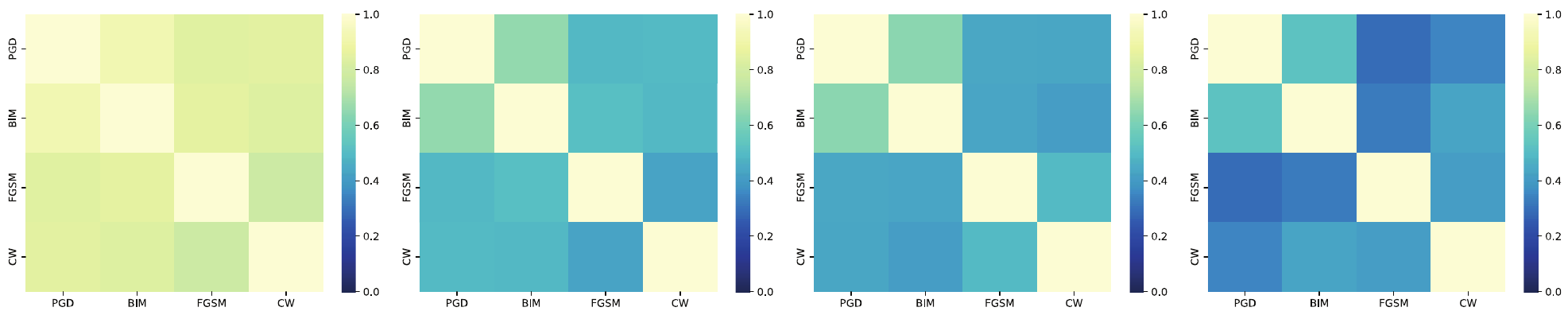}
            \caption{Untargeted Attacks.}
    \end{subfigure}
    \caption{The similarity of different adversarial attacks. }
\label{Fig:adv_target}
\end{figure*}

In the end, we use GBS to reveal the transferability of adversarial attacks in representations and the difference between targeted and untargeted attacks. To analyze whether the internal representations are similar when the attack transfers to different models, we generate adversarial samples against one model and attack other models with them. The models used here are VGGNets of different initialization. We use \textit{advertorch}~\cite{ding2019advertorch} package to generate untargeted adversarial samples for the VGGNet16 model trained on CIFAR10. The adversarial attacks include Fast Gradient Sign Method~\cite{goodfellow2014explaining} (FGSM), Carlini \& Wagner Attack~\cite{carlini2017towards} (CW), Basic Iterative Method~\cite{kurakin2016adversarial} (BIM), and Project Gradient Descent~\cite{madry2017towards} (PGD). All the attacks use the default parameters and the attack success rates of PGD, BIM, FGSM, and CW for the original model are 56\%, 50\%, 41\%, and 54\%, respectively. The batch size is set to 100. As shown in \autoref{Fig:trans} (a), compared with the normal samples, all attacks show that the similarity decreases with the increase of depth. When the same batch of normal samples are fed into other models, as shown in \autoref{Fig:trans}, the representations show relatively high similarity at all depths. However, when fed with the same batch of adversarial samples, the adversarial samples successfully fool them with different representations activated in the other three models. It indicates that the transferability of adversarial samples is not because the same features are activated.

In \autoref{Fig:adv_target}, we show the similarity between different types of adversarial examples in the hidden layer space. We calculate the similarity between different adversarial attacks every 3 layers in VGGNet16. Targeted attacks and untargeted attacks are calculated separately and the target label is all set to the same one. For both kinds of attacks, shallow layers have a high similarity between different attacks, and their representations diverge with the depth going deeper. Against our intuition that targeted attacks should be more similar than untargeted attacks because the model classifies them into the same category, untargeted attacks show higher similarity than targeted attacks in deep layers. We argue that since the object of the untargeted attack is only decreasing the original output confidence, it weakens the representations of the original object. The targeted attack aims to increase the confidence in a certain category, thus it brings new representations of the target category, weakening the original representation at the same time. Therefore, the similarity between targeted attacks is lower than untargeted ones, since they have multiple representations in the hidden layer space.

\section{Conclusion}
\label{Sec:Conclusion}
In this paper, we first revisit CKA from a graph perspective, then propose Graph-Based Similarity (GBS) to measure the similarity between representations of DNN. In addition to representational similarity, the ratio of type1 motifs in the last layer is closely related to the functional similarity of DNN. By sanity check, we show that GBS is a more fine-grained representational similarity index than CKA, which is robust to changes in graph size and degree. We compared the similarity of representations when the same batch of adversarial samples migrated to different models and found that the same adversarial samples activated different features in different models. The similarity between representations of untargeted adversarial attacks is higher than that of targeted attacks.
\clearpage

\bibliographystyle{plain}

\begin{thebibliography}{10}

\bibitem{abu2015exact}
Zeina Abu-Aisheh, Romain Raveaux, Jean-Yves Ramel, and Patrick Martineau.
\newblock An exact graph edit distance algorithm for solving pattern
  recognition problems.
\newblock In {\em 4th International Conference on Pattern Recognition
  Applications and Methods 2015}, 2015.

\bibitem{abusnaina2021adversarial}
Ahmed Abusnaina, Yuhang Wu, Sunpreet Arora, Yizhen Wang, Fei Wang, Hao Yang,
  and David Mohaisen.
\newblock Adversarial example detection using latent neighborhood graph.
\newblock In {\em Proceedings of the IEEE/CVF International Conference on
  Computer Vision}, pages 7687--7696, 2021.

\bibitem{akhtar2018threat}
Naveed Akhtar and Ajmal Mian.
\newblock Threat of adversarial attacks on deep learning in computer vision: A
  survey.
\newblock {\em IEEE Access}, 6:14410--14430, 2018.

\bibitem{bach2015pixel}
Sebastian Bach, Alexander Binder, Gr{\'e}goire Montavon, Frederick Klauschen,
  Klaus-Robert M{\"u}ller, and Wojciech Samek.
\newblock On pixel-wise explanations for non-linear classifier decisions by
  layer-wise relevance propagation.
\newblock {\em PloS one}, 10(7):e0130140, 2015.

\bibitem{baker1973joint}
Charles~R Baker.
\newblock Joint measures and cross-covariance operators.
\newblock {\em Transactions of the American Mathematical Society},
  186:273--289, 1973.

\bibitem{bansal2021revisiting}
Yamini Bansal, Preetum Nakkiran, and Boaz Barak.
\newblock Revisiting model stitching to compare neural representations.
\newblock {\em Advances in Neural Information Processing Systems}, 34, 2021.

\bibitem{berretti2001efficient}
Stefano Berretti, Alberto Del~Bimbo, and Enrico Vicario.
\newblock Efficient matching and indexing of graph models in content-based
  retrieval.
\newblock {\em IEEE Transactions on Pattern Analysis and Machine Intelligence},
  23(10):1089--1105, 2001.

\bibitem{boccaletti2014structure}
Stefano Boccaletti, Ginestra Bianconi, Regino Criado, Charo~I Del~Genio,
  Jes{\'u}s G{\'o}mez-Gardenes, Miguel Romance, Irene Sendina-Nadal, Zhen Wang,
  and Massimiliano Zanin.
\newblock The structure and dynamics of multilayer networks.
\newblock {\em Physics reports}, 544(1):1--122, 2014.

\bibitem{bolya2021scalable}
Daniel Bolya, Rohit Mittapalli, and Judy Hoffman.
\newblock Scalable diverse model selection for accessible transfer learning.
\newblock {\em NIPS}, 34, 2021.

\bibitem{brown2020language}
Tom~B Brown, Benjamin Mann, Nick Ryder, Melanie Subbiah, Jared Kaplan, Prafulla
  Dhariwal, Arvind Neelakantan, Pranav Shyam, Girish Sastry, Amanda Askell,
  et~al.
\newblock Language models are few-shot learners.
\newblock {\em arXiv preprint arXiv:2005.14165}, 2020.

\bibitem{carlini2017towards}
Nicholas Carlini and David Wagner.
\newblock Towards evaluating the robustness of neural networks.
\newblock In {\em IEEE Symposium on Security and Privacy}, pages 39--57, 2017.

\bibitem{caron2018deep}
Mathilde Caron, Piotr Bojanowski, Armand Joulin, and Matthijs Douze.
\newblock Deep clustering for unsupervised learning of visual features.
\newblock In {\em Proceedings of the European conference on computer vision
  (ECCV)}, pages 132--149, 2018.

\bibitem{cortes2012algorithms}
Corinna Cortes, Mehryar Mohri, and Afshin Rostamizadeh.
\newblock Algorithms for learning kernels based on centered alignment.
\newblock {\em The J. of Machine Learning Research}, 13(1):795--828, 2012.

\bibitem{csiszarik2021similarity}
Adri{\'a}n Csisz{\'a}rik, P{\'e}ter K{\H{o}}r{\"o}si-Szab{\'o}, {\'A}kos
  Matszangosz, Gergely Papp, and D{\'a}niel Varga.
\newblock Similarity and matching of neural network representations.
\newblock {\em Advances in Neural Information Processing Systems}, 34, 2021.

\bibitem{cui2022deconfounded}
Tianyu Cui, Yogesh Kumar, Pekka Marttinen, and Samuel Kaski.
\newblock Deconfounded representation similarity for comparison of neural
  networks.
\newblock {\em arXiv preprint arXiv:2202.00095}, 2022.

\bibitem{das2020opportunities}
Arun Das and Paul Rad.
\newblock Opportunities and challenges in explainable artificial intelligence
  (xai): A survey.
\newblock {\em arXiv preprint arXiv:2006.11371}, 2020.

\bibitem{5206848}
Jia Deng, Wei Dong, Richard Socher, Li-Jia Li, Kai Li, and Li~Fei-Fei.
\newblock Imagenet:{ A }large-scale hierarchical image database.
\newblock In {\em CVPR}, pages 248--255, 2009.

\bibitem{ding2021grounding}
Frances Ding, Jean-Stanislas Denain, and Jacob Steinhardt.
\newblock Grounding representation similarity through statistical testing.
\newblock In {\em Advances in Neural Information Processing Systems}, 2021.

\bibitem{ding2019advertorch}
Gavin~Weiguang Ding, Luyu Wang, and Xiaomeng Jin.
\newblock {AdverTorch} v0.1: An adversarial robustness toolbox based on
  pytorch.
\newblock {\em arXiv preprint arXiv:1902.07623}, 2019.

\bibitem{dosovitskiy2020image}
Alexey Dosovitskiy, Lucas Beyer, Alexander Kolesnikov, Dirk Weissenborn,
  Xiaohua Zhai, Thomas Unterthiner, Mostafa Dehghani, Matthias Minderer, Georg
  Heigold, Sylvain Gelly, et~al.
\newblock An image is worth 16x16 words: Transformers for image recognition at
  scale.
\newblock {\em arXiv preprint arXiv:2010.11929}, 2020.

\bibitem{dwivedi2019representation}
Kshitij Dwivedi and Gemma Roig.
\newblock Representation similarity analysis for efficient task taxonomy \&
  transfer learning.
\newblock In {\em CVPR}, pages 12387--12396, 2019.

\bibitem{filan2021clusterability}
Daniel Filan, Stephen Casper, Shlomi Hod, Cody Wild, Andrew Critch, and Stuart
  Russell.
\newblock Clusterability in neural networks.
\newblock {\em arXiv preprint arXiv:2103.03386}, 2021.

\bibitem{frankle2018lottery}
Jonathan Frankle and Michael Carbin.
\newblock The lottery ticket hypothesis: Finding sparse, trainable neural
  networks.
\newblock {\em arXiv preprint arXiv:1803.03635}, 2018.

\bibitem{fukumizu2004dimensionality}
Kenji Fukumizu, Francis~R Bach, and Michael~I Jordan.
\newblock Dimensionality reduction for supervised learning with reproducing
  kernel hilbert spaces.
\newblock {\em Journal of Machine Learning Research}, 5(Jan):73--99, 2004.

\bibitem{ge2021peek}
Yunhao Ge, Yao Xiao, Zhi Xu, Meng Zheng, Srikrishna Karanam, Terrence Chen,
  Laurent Itti, and Ziyan Wu.
\newblock A peek into the reasoning of neural networks: Interpreting with
  structural visual concepts.
\newblock In {\em CVPR}, pages 2195--2204, 2021.

\bibitem{glorot2010understanding}
Xavier Glorot and Yoshua Bengio.
\newblock Understanding the difficulty of training deep feedforward neural
  networks.
\newblock In {\em Proc. of the Int. Conf. on Artificial Intelligence and
  Statistics}, pages 249--256, 2010.

\bibitem{goodfellow2014explaining}
Ian~J Goodfellow, Jonathon Shlens, and Christian Szegedy.
\newblock Explaining and harnessing adversarial examples.
\newblock {\em arXiv preprint arXiv:1412.6572}, 2014.

\bibitem{goyal2019explaining}
Yash Goyal, Amir Feder, Uri Shalit, and Been Kim.
\newblock Explaining classifiers with causal concept effect (cace).
\newblock {\em arXiv preprint arXiv:1907.07165}, 2019.

\bibitem{gretton2005measuring}
Arthur Gretton, Olivier Bousquet, Alex Smola, and Bernhard Sch{\"o}lkopf.
\newblock Measuring statistical dependence with hilbert-schmidt norms.
\newblock In {\em International conference on algorithmic learning theory},
  pages 63--77. Springer, 2005.

\bibitem{guo2020backpropagating}
Yiwen Guo, Qizhang Li, and Hao Chen.
\newblock Backpropagating linearly improves transferability of adversarial
  examples.
\newblock {\em arXiv preprint arXiv:2012.03528}, 2020.

\bibitem{hardoon2004canonical}
David~R Hardoon, Sandor Szedmak, and John Shawe-Taylor.
\newblock Canonical correlation analysis: An overview with application to
  learning methods.
\newblock {\em Neural Computation}, 16(12):2639--2664, 2004.

\bibitem{he2015delving}
Kaiming He, Xiangyu Zhang, Shaoqing Ren, and Jian Sun.
\newblock Delving deep into rectifiers: Surpassing human-level performance on
  imagenet classification.
\newblock In {\em ICCV}, pages 1026--1034, 2015.

\bibitem{he2016deep}
Kaiming He, Xiangyu Zhang, Shaoqing Ren, and Jian Sun.
\newblock Deep residual learning for image recognition.
\newblock In {\em CVPR}, pages 770--778, 2016.

\bibitem{horta2021extracting}
Vitor~AC Horta, Ilaria Tiddi, Suzanne Little, and Alessandra Mileo.
\newblock Extracting knowledge from deep neural networks through graph
  analysis.
\newblock {\em Future Generation Computer Systems}, 120:109--118, 2021.

\bibitem{ilyas2019adversarial}
Andrew Ilyas, Shibani Santurkar, Dimitris Tsipras, Logan Engstrom, Brandon
  Tran, and Aleksander Madry.
\newblock Adversarial examples are not bugs, they are features.
\newblock In {\em Advances in Neural Information Processing Systems}, pages
  125--136, 2019.

\bibitem{ioffe2015batch}
Sergey Ioffe and Christian Szegedy.
\newblock Batch normalization: Accelerating deep network training by reducing
  internal covariate shift.
\newblock In {\em ICML}, pages 448--456. PMLR, 2015.

\bibitem{jeh2002simrank}
Glen Jeh and Jennifer Widom.
\newblock Simrank: a measure of structural-context similarity.
\newblock In {\em Proceedings of the eighth ACM SIGKDD international conference
  on Knowledge discovery and data mining}, pages 538--543, 2002.

\bibitem{kornblith2019similarity}
Simon Kornblith, Mohammad Norouzi, Honglak Lee, and Geoffrey Hinton.
\newblock Similarity of neural network representations revisited.
\newblock In {\em ICML}, pages 3519--3529. PMLR, 2019.

\bibitem{krizhevsky2009learning}
Alex Krizhevsky, Geoffrey Hinton, et~al.
\newblock Learning multiple layers of features from tiny images.
\newblock 2009.

\bibitem{kurakin2016adversarial}
Alexey Kurakin, Ian Goodfellow, Samy Bengio, et~al.
\newblock Adversarial examples in the physical world, 2016.

\bibitem{lee2017deep}
Jaehoon Lee, Yasaman Bahri, Roman Novak, Samuel~S Schoenholz, Jeffrey
  Pennington, and Jascha Sohl-Dickstein.
\newblock Deep neural networks as gaussian processes.
\newblock {\em arXiv preprint arXiv:1711.00165}, 2017.

\bibitem{liu2019knowledge}
Yufan Liu, Jiajiong Cao, Bing Li, Chunfeng Yuan, Weiming Hu, Yangxi Li, and
  Yunqiang Duan.
\newblock Knowledge distillation via instance relationship graph.
\newblock In {\em Proceedings of the IEEE/CVF Conference on Computer Vision and
  Pattern Recognition}, pages 7096--7104, 2019.

\bibitem{madry2017towards}
Aleksander Madry, Aleksandar Makelov, Ludwig Schmidt, Dimitris Tsipras, and
  Adrian Vladu.
\newblock Towards deep learning models resistant to adversarial attacks.
\newblock {\em arXiv preprint arXiv:1706.06083}, 2017.

\bibitem{mahendran2015understanding}
Aravindh Mahendran and Andrea Vedaldi.
\newblock Understanding deep image representations by inverting them.
\newblock In {\em CVPR}, pages 5188--5196, 2015.

\bibitem{mehrer2020individual}
Johannes Mehrer, Courtney~J Spoerer, Nikolaus Kriegeskorte, and Tim~C
  Kietzmann.
\newblock Individual differences among deep neural network models.
\newblock {\em Nature Communications}, 11(1):1--12, 2020.

\bibitem{montavon2018methods}
Gr{\'e}goire Montavon, Wojciech Samek, and Klaus-Robert M{\"u}ller.
\newblock Methods for interpreting and understanding deep neural networks.
\newblock {\em Digital Signal Processing}, 73:1--15, 2018.

\bibitem{morcos2018insights}
Ari~S Morcos, Maithra Raghu, and Samy Bengio.
\newblock Insights on representational similarity in neural networks with
  canonical correlation.
\newblock {\em arXiv preprint arXiv:1806.05759}, 2018.

\bibitem{muzellec2018generalizing}
Boris Muzellec and Marco Cuturi.
\newblock Generalizing point embeddings using the wasserstein space of
  elliptical distributions.
\newblock In {\em NIPS}, pages 10258--10269, 2018.

\bibitem{nguyen2020wide}
Thao Nguyen, Maithra Raghu, and Simon Kornblith.
\newblock Do wide and deep networks learn the same things? uncovering how
  neural network representations vary with width and depth.
\newblock {\em arXiv preprint arXiv:2010.15327}, 2020.

\bibitem{nguyen2022origins}
Thao Nguyen, Maithra Raghu, and Simon Kornblith.
\newblock On the origins of the block structure phenomenon in neural network
  representations.
\newblock {\em arXiv preprint arXiv:2202.07184}, 2022.

\bibitem{orhan2017skip}
A~Emin Orhan and Xaq Pitkow.
\newblock Skip connections eliminate singularities.
\newblock {\em arXiv preprint arXiv:1701.09175}, 2017.

\bibitem{qin2020binary}
Haotong Qin, Ruihao Gong, Xianglong Liu, Xiao Bai, Jingkuan Song, and Nicu
  Sebe.
\newblock Binary neural networks: A survey.
\newblock {\em Pattern Recognition}, 105:107281, 2020.

\bibitem{raghu2017svcca}
Maithra Raghu, Justin Gilmer, Jason Yosinski, and Jascha Sohl-Dickstein.
\newblock Svcca: Singular vector canonical correlation analysis for deep
  learning dynamics and interpretability.
\newblock {\em arXiv preprint arXiv:1706.05806}, 2017.

\bibitem{raghu2021vision}
Maithra Raghu, Thomas Unterthiner, Simon Kornblith, Chiyuan Zhang, and Alexey
  Dosovitskiy.
\newblock Do vision transformers see like convolutional neural networks?
\newblock {\em arXiv preprint arXiv:2108.08810}, 2021.

\bibitem{ramanujan2020s}
Vivek Ramanujan, Mitchell Wortsman, Aniruddha Kembhavi, Ali Farhadi, and
  Mohammad Rastegari.
\newblock What's hidden in a randomly weighted neural network?
\newblock In {\em CVPR}, pages 11893--11902, 2020.

\bibitem{ranftl2021vision}
Ren{\'e} Ranftl, Alexey Bochkovskiy, and Vladlen Koltun.
\newblock Vision transformers for dense prediction.
\newblock In {\em CVPR}, pages 12179--12188, 2021.

\bibitem{raymond2002rascal}
John~W Raymond, Eleanor~J Gardiner, and Peter Willett.
\newblock Rascal: Calculation of graph similarity using maximum common edge
  subgraphs.
\newblock {\em The Computer Journal}, 45(6):631--644, 2002.

\bibitem{selvaraju2017grad}
Ramprasaath~R Selvaraju, Michael Cogswell, Abhishek Das, Ramakrishna Vedantam,
  Devi Parikh, and Dhruv Batra.
\newblock Grad-cam: {Visual} explanations from deep networks via gradient-based
  localization.
\newblock In {\em ICCV}, pages 618--626, 2017.

\bibitem{simonyan2013deep}
Karen Simonyan, Andrea Vedaldi, and Andrew Zisserman.
\newblock Deep inside convolutional networks: Visualising image classification
  models and saliency maps.
\newblock {\em arXiv preprint arXiv:1312.6034}, 2013.

\bibitem{simonyan2014very}
Karen Simonyan and Andrew Zisserman.
\newblock Very deep convolutional networks for large-scale image recognition.
\newblock {\em arXiv preprint arXiv:1409.1556}, 2014.

\bibitem{song2020depara}
Jie Song, Yixin Chen, Jingwen Ye, Xinchao Wang, Chengchao Shen, Feng Mao, and
  Mingli Song.
\newblock Depara: {Deep} attribution graph for deep knowledge transferability.
\newblock In {\em CVPR}, pages 3922--3930, 2020.

\bibitem{springenberg2014striving}
Jost~Tobias Springenberg, Alexey Dosovitskiy, Thomas Brox, and Martin
  Riedmiller.
\newblock Striving for simplicity: The all convolutional net.
\newblock {\em arXiv preprint arXiv:1412.6806}, 2014.

\bibitem{sung2018learning}
Flood Sung, Yongxin Yang, Li~Zhang, Tao Xiang, Philip~HS Torr, and Timothy~M
  Hospedales.
\newblock Learning to compare: Relation network for few-shot learning.
\newblock In {\em Proceedings of the IEEE conference on computer vision and
  pattern recognition}, pages 1199--1208, 2018.

\bibitem{szegedy2013intriguing}
Christian Szegedy, Wojciech Zaremba, Ilya Sutskever, Joan Bruna, Dumitru Erhan,
  Ian Goodfellow, and Rob Fergus.
\newblock Intriguing properties of neural networks.
\newblock {\em arXiv preprint arXiv:1312.6199}, 2013.

\bibitem{tang2020similarity}
Shuai Tang, Wesley~J Maddox, Charlie Dickens, Tom Diethe, and Andreas Damianou.
\newblock Similarity of neural networks with gradients.
\newblock {\em arXiv preprint arXiv:2003.11498}, 2020.

\bibitem{tolstikhin2021mlp}
Ilya Tolstikhin, Neil Houlsby, Alexander Kolesnikov, Lucas Beyer, Xiaohua Zhai,
  Thomas Unterthiner, Jessica Yung, Andreas Steiner, Daniel Keysers, Jakob
  Uszkoreit, et~al.
\newblock Mlp-mixer: An all-mlp architecture for vision.
\newblock {\em arXiv preprint arXiv:2105.01601}, 2021.

\bibitem{wang2018interpret}
Yulong Wang, Hang Su, Bo~Zhang, and Xiaolin Hu.
\newblock Interpret neural networks by identifying critical data routing paths.
\newblock In {\em CVPR}, pages 8906--8914, 2018.

\bibitem{wang2019learning}
Yulong Wang, Hang Su, Bo~Zhang, and Xiaolin Hu.
\newblock Learning reliable visual saliency for model explanations.
\newblock {\em TMM}, 22(7):1796--1807, 2019.

\bibitem{watanabe2019understanding}
Chihiro Watanabe, Kaoru Hiramatsu, and Kunio Kashino.
\newblock Understanding community structure in layered neural networks.
\newblock {\em Neurocomputing}, 367:84--102, 2019.

\bibitem{watts1998collective}
Duncan~J Watts and Steven~H Strogatz.
\newblock Collective dynamics of ‘small-world’networks.
\newblock {\em nature}, 393(6684):440--442, 1998.

\bibitem{willett1998chemical}
Peter Willett, John~M Barnard, and Geoffrey~M Downs.
\newblock Chemical similarity searching.
\newblock {\em Journal of chemical information and computer sciences},
  38(6):983--996, 1998.

\bibitem{williams2021generalized}
Alex Williams, Erin Kunz, Simon Kornblith, and Scott Linderman.
\newblock Generalized shape metrics on neural representations.
\newblock {\em Advances in Neural Information Processing Systems}, 34, 2021.

\bibitem{yan2004graph}
Xifeng Yan, Philip~S Yu, and Jiawei Han.
\newblock Graph indexing: a frequent structure-based approach.
\newblock In {\em Proceedings of the 2004 ACM SIGMOD international conference
  on Management of data}, pages 335--346, 2004.

\bibitem{yosinski2014transferable}
Jason Yosinski, Jeff Clune, Yoshua Bengio, and Hod Lipson.
\newblock How transferable are features in deep neural networks?
\newblock {\em arXiv preprint arXiv:1411.1792}, 2014.

\bibitem{you2020graph}
Jiaxuan You, Jure Leskovec, Kaiming He, and Saining Xie.
\newblock Graph structure of neural networks.
\newblock In {\em ICML}, pages 10881--10891. PMLR, 2020.

\bibitem{yuan2020interpreting}
Hao Yuan, Lei Cai, Xia Hu, Jie Wang, and Shuiwang Ji.
\newblock Interpreting image classifiers by generating discrete masks.
\newblock {\em PAMI}, 2020.

\bibitem{zeiler2014visualizing}
Matthew~D Zeiler and Rob Fergus.
\newblock Visualizing and understanding convolutional networks.
\newblock In {\em ECCV}, pages 818--833. Springer, 2014.

\bibitem{zhang2020understanding}
Chaoning Zhang, Philipp Benz, Tooba Imtiaz, and In~So Kweon.
\newblock Understanding adversarial examples from the mutual influence of
  images and perturbations.
\newblock In {\em CVPR}, pages 14521--14530, 2020.

\bibitem{zhang2020interpreting}
Chongzhi Zhang, Aishan Liu, Xianglong Liu, Yitao Xu, Hang Yu, Yuqing Ma, and
  Tianlin Li.
\newblock Interpreting and improving adversarial robustness of deep neural
  networks with neuron sensitivity.
\newblock {\em IEEE TIP}, 30:1291--1304, 2020.

\bibitem{zhang2020measuring}
Ronda~J Zhang and Y~Ye Fred.
\newblock Measuring similarity for clarifying layer difference in multiplex ad
  hoc duplex information networks.
\newblock {\em J. of Informetrics}, 14(1):100987, 2020.

\bibitem{zhou2016normalization}
Qiuju Zhou and Loet Leydesdorff.
\newblock The normalization of occurrence and c o-occurrence matrices in
  bibliometrics using cosine similarities and o chiai coefficients.
\newblock {\em J. of the Association for Information Science and Technology},
  67(11):2805--2814, 2016.

\bibitem{zhou2021distilling}
Sheng Zhou, Yucheng Wang, Defang Chen, Jiawei Chen, Xin Wang, Can Wang, and
  Jiajun Bu.
\newblock Distilling holistic knowledge with graph neural networks.
\newblock In {\em CVPR}, pages 10387--10396, 2021.

\end{thebibliography}

\clearpage

\appendix

\setcounter{page}{1}







\section{Invariance Property of GBS}
GBS first constructs the representation graph, then calculates the similarity between graphs. The graph $\mathcal{G}_i=(\mathcal{V}, \mathcal{E})$ is defined by its nodes and edges. We use the adjacent matrix $\textbf{A}_i$ and $\textbf{A}_j$ to calculate the similarity
\begin{equation}
s_{ij} = \frac{1}{N}\sum_{k=1}^N f_{cos}(\textbf{a}_{i,k}, \textbf{a}_{j,k}),
\label{eq:gsimilarity}
\end{equation}
where $\textbf{a}_{i,k}$ is the row vector of $\textbf{A}_i$ and $f_{cos}(\cdot)$ is the cosine similarity between two vectors.
\autoref{eq:gsimilarity} is correlated with the number of nodes $N$ and the edge weights. $N$ only depends on the number of inputs, thus edge weights are the only factor that affects GBS. In the step of constructing graphs, edge weights are computed by 
\begin{equation}
    \mathit{e}_{nm} = \dfrac{\textbf{F}_{i|n} \cdot \textbf{F}^T_{i|m}}{\Vert \textbf{F}_{i|n} \Vert _2 \cdot \Vert \textbf{F}_{i|m} \Vert _2}.
\end{equation}

GBS is invariant to orthogonal transformation (OT) and isotropic scaling (IS) but not invertible linear transformation (ILT). The edge weight after ILT is 
\begin{equation}
    ILT(\mathit{e}_{nm}) = \dfrac{\textbf{B}\textbf{F}_{i|n} \cdot \textbf{F}^T_{i|m}}{\Vert \textbf{B}\textbf{F}_{i|n} \Vert _2 \cdot \Vert \textbf{F}_{i|m} \Vert _2},
    \label{eq:cos}
\end{equation}
where $\textbf{B}$ is an arbitrary full rank matrix with the same size of $\textbf{F}$. 

According to the definition of OT, it preserves symmetric inner product, i.e., preserving lengths of vectors and angles between vectors. Then for two vectors $\textbf{u}$ and $\textbf{v}$
\begin{equation}
\begin{split}
     ||OT(\textbf{v})||  & = ||\textbf{v}|| \\
   <\textbf{u}, \textbf{v}> &  = <OT(\textbf{u}), OT(\textbf{v})>.
\end{split}
\end{equation}
The edges are weighted with cosine similarity using \autoref{eq:cos}, which is only relevant to the angles between the two vectors. Since OT does not change the angles between vectors, GBS is invariant to OT.

IS is the rescaling of representations, i.e., the same scales are applied to each dimension. After IS, a vector $\textbf{v}$ can be expressed as $\alpha \textbf{v}$, where $\alpha \in \mathbb{R}^+$. The weight of the edge between two nodes is
\begin{equation}
    IS(\mathit{e}_{nm}) = \dfrac{\alpha\textbf{F}_{i|n} \cdot \textbf{F}^T_{i|m}}{\Vert \alpha\textbf{F}_{i|n} \Vert _2 \cdot \Vert \textbf{F}_{i|m} \Vert _2}.
\end{equation}
Since the graph edges are not affected by IS, GBS is invariant to IS.


\section{Network Structure}
The detailed network structures of ResNets, VGGNets, and 10-layer CNN used in our experiments are listed in \autoref{Tab:resarch}, \autoref{Tab:vggarch}, and \autoref{Tab:10cnnarch}.

\begin{table}[!h]
	\caption{VGGNets Configuration.}
	\label{Tab:vggarch}
    \centering
	\begin{tabular}{cccc}
	\hline
	  VGG11 & VGG13 & VGG16 & VGG19  \tabularnewline
		\hline
          conv3-64 &  conv3-64$\times$2  & conv3-64$\times$2  &   conv3-64$\times$2   \tabularnewline
		  \hline
          conv3-128 & conv3-128$\times$2  & conv3-128$\times$2 &  conv3-128$\times$2 \tabularnewline
         \hline
          conv3-256$\times$2  & conv3-256$\times$2 & conv3-256$\times$3 &  conv3-256$\times$4  \tabularnewline
         \hline
            conv3-512$\times$2 & conv3-512$\times$2 & conv3-512$\times$3 &  conv3-512$\times$4 \tabularnewline   
          \hline
          conv3-512$\times$2 & conv3-512$\times$2 & conv3-512$\times$3 &  conv3-512$\times$4  \tabularnewline   
          \hline
	\end{tabular}
\end{table}

\begin{table*}[t]\footnotesize
  \centering
  \caption{ResNets Configuration.}

  \resizebox{0.9\textwidth}{!}{
    \begin{tabular}{c p{0.5cm} c p{0.5cm} c p{0.5cm} c p{0.5cm} c p{0.5cm} }
\hline
\multicolumn{2}{c}{ResNet18 } & \multicolumn{2}{c}{ResNet34} & \multicolumn{2}{c}{ResNet50} & \multicolumn{2}{c}{ResNet101} & \multicolumn{2}{c}{ResNet152} \tabularnewline
    \hline 
    conv3-64 & \multirow{3}[0]{*}{$\times$2} & {conv3-64} & \multirow{3}[0]{*}{$\times$3} & conv1-64 & \multirow{3}[0]{*}{$\times$3} & conv1-64 & \multirow{3}[0]{*}{$\times$3} & conv1-64 & \multirow{3}[0]{*}{$\times$3} \tabularnewline

    {conv3-64} & ~ & {conv3-64} & ~ & conv3-64 & ~ & conv3-64 & ~ & conv3-64 &  \tabularnewline

    &&&& conv1-256 && conv1-256 && conv1-256 &  \tabularnewline
    \hline
    {conv3-128} & \multirow{3}[0]{*}{$\times$2} & {conv3-128} & \multirow{3}[0]{*}{$\times$4} & conv1-128 & \multirow{3}[0]{*}{$\times$4} & conv1-128 & \multirow{3}[0]{*}{$\times$4} & conv1-128 & \multirow{3}[0]{*}{$\times$8} \tabularnewline

    {conv3-128} && {conv3-128} & & conv3-128 & & conv3-128 & & conv3-128 &  \tabularnewline

    &&&& conv1-512 & & conv1-512 & & conv1-512 &  \tabularnewline
    \hline
    {conv3-256} & \multirow{3}[0]{*}{$\times$2} & {conv3-256} & \multirow{3}[0]{*}{$\times$6} & conv1-256 & \multirow{3}[0]{*}{$\times$6} & conv1-256 & \multirow{3}[0]{*}{$\times$23} & conv1-256 & \multirow{3}[0]{*}{$\times$36} \tabularnewline
    	
    {conv3-256} && {conv3-256} & & conv3-256 & & conv3-256 & & conv3-256 &  \tabularnewline
    	
    &&&& conv1-1024 & & conv1-1024 & & conv1-1024 &  \tabularnewline
    	\hline
    {conv3-512} & \multirow{3}[0]{*}{$\times$2} & {conv3-512} & \multirow{3}[0]{*}{$\times$3} & conv1-512 & \multirow{3}[0]{*}{$\times$3} & conv1-512 & \multirow{3}[0]{*}{$\times$3} & conv1-512 & \multirow{3}[0]{*}{$\times$3} \tabularnewline
    	
    {conv3-512} && {conv3-512} & & conv3-512 & & conv3-512 & & conv3-512 &  \tabularnewline
    	
    &&&& conv1-2048 && conv1-2048 && conv1-2048 &  \tabularnewline
    	\hline
    \end{tabular}%
  }
  \label{Tab:resarch}%
\end{table*}%

\begin{table}[!h]
	\caption{10-layer CNN Configuration.}
	\label{Tab:10cnnarch}
    \centering
	\begin{tabular}{c}
	\hline
	  10-layer CNN  \tabularnewline
		\hline
          conv3-16 BN ReLu$\times$2  \tabularnewline
		  \hline
          conv3-32 (stride=2)BN ReLu \tabularnewline
		  \hline
		  conv3-32 BN ReLu $\times$2  \tabularnewline
		  \hline
		  conv3-64 (stride=2)BN ReLu  \tabularnewline
		  \hline
		  conv3-64 (valid padding) \tabularnewline
		  \hline
		  conv1-64 BN ReLu  \tabularnewline
		  \hline
		  Global average pooling  \tabularnewline
          \hline
	\end{tabular}
\end{table}

\section{Sanity Check Results}
\begin{figure*}[!t]
    \centering
    \begin{subfigure}[b]{0.95\textwidth}
           \centering
           \includegraphics[width=\textwidth]{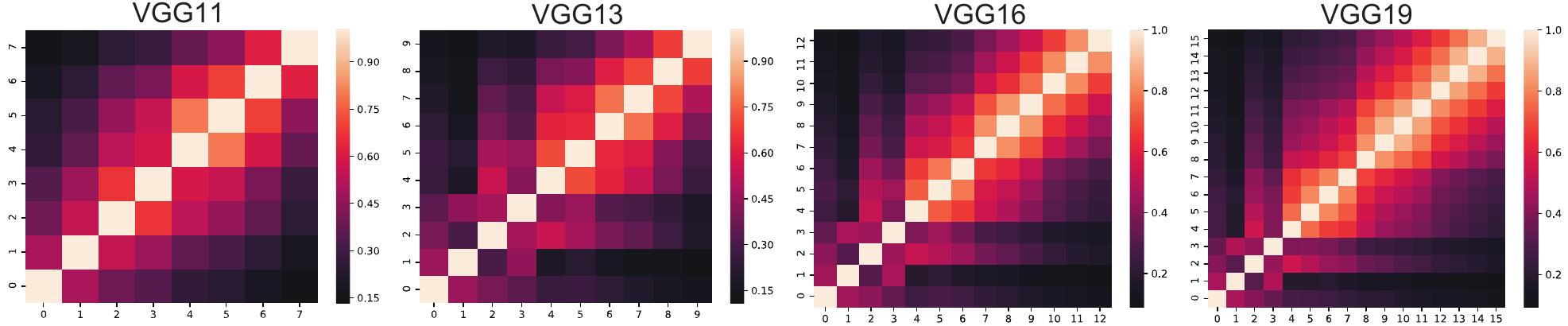}
    \end{subfigure}
    \begin{subfigure}[b]{0.95\textwidth}
            \centering
            \includegraphics[width=\textwidth]{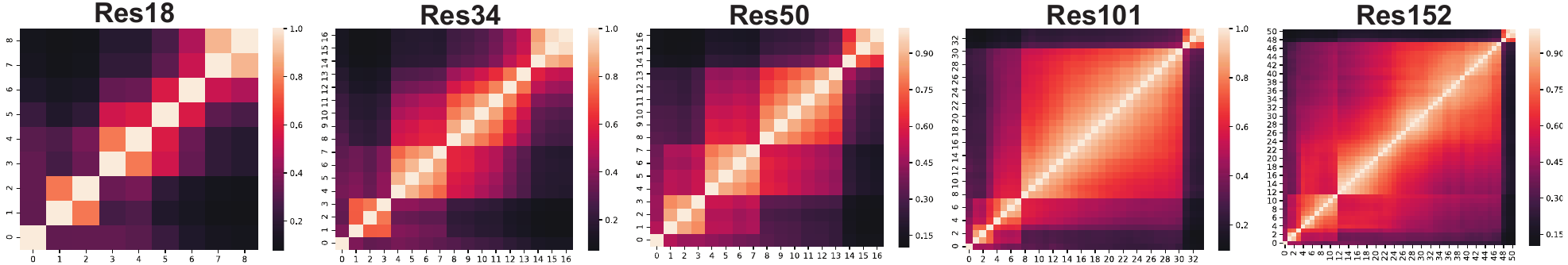}
         
    \end{subfigure}
    \caption{The similarity between the layers of VGGNets and ResNets on CIFAR100.}
    \label{Fig:path_c100}
\end{figure*}

\begin{figure*}[!t]
    \centering
    \begin{subfigure}[b]{0.95\textwidth}
           \centering
           \includegraphics[width=\textwidth]{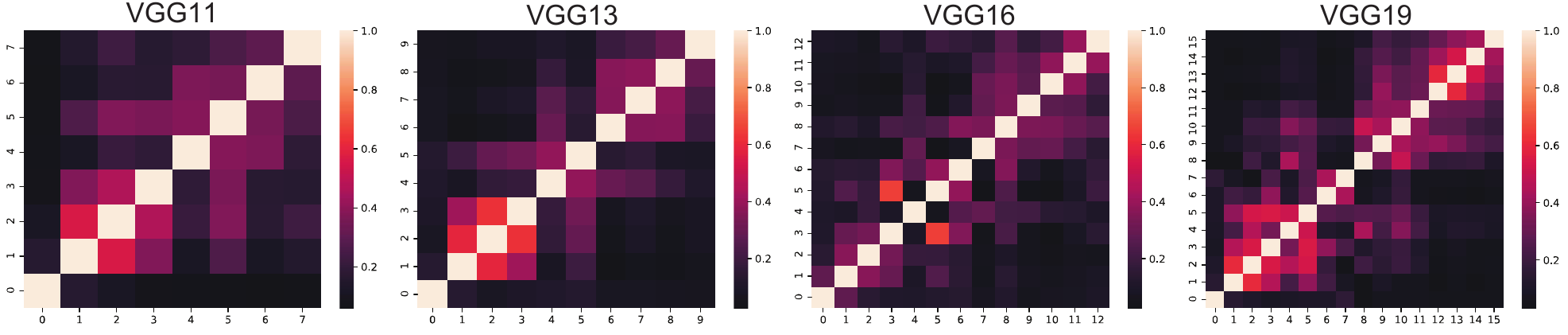}
    \end{subfigure}
    \begin{subfigure}[b]{0.95\textwidth}
            \centering
            \includegraphics[width=\textwidth]{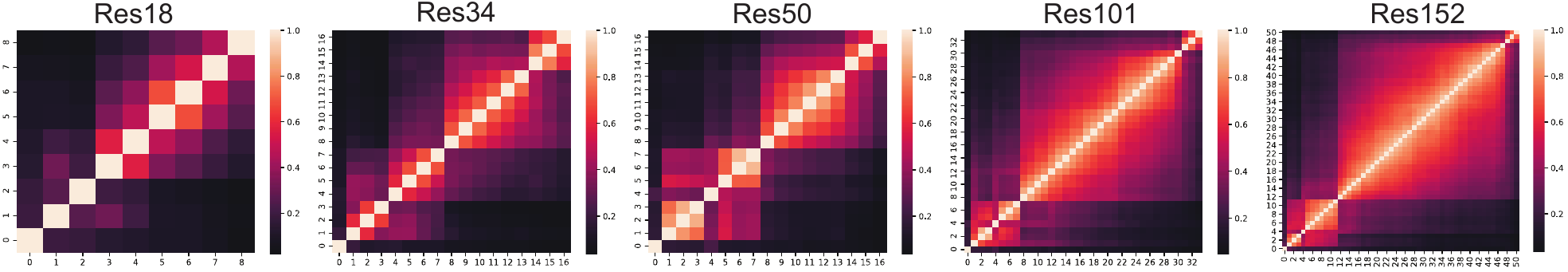}
         
    \end{subfigure}
    \caption{The similarity between the layers of VGGNets and ResNets on ImageNet.}
    \label{Fig:path_imnt}
\end{figure*}

The network pathology of VGGNets and ResNets on CIFAR100 and ImageNet are shown in \autoref{Fig:path_c100} and \autoref{Fig:path_imnt}. The pathology of the model on different data is also different. On CIFAR100, the adjacent grids of ResNets show higher similarity between each other, while they are less similar on ImageNet. This phenomenon is more clear in VGGNets. For VGGNets, the similarity is very low even within a block on ImageNet.

\begin{figure*}[!t]
    \centering
    \begin{subfigure}[b]{0.95\textwidth}
           \centering
           \includegraphics[width=\textwidth]{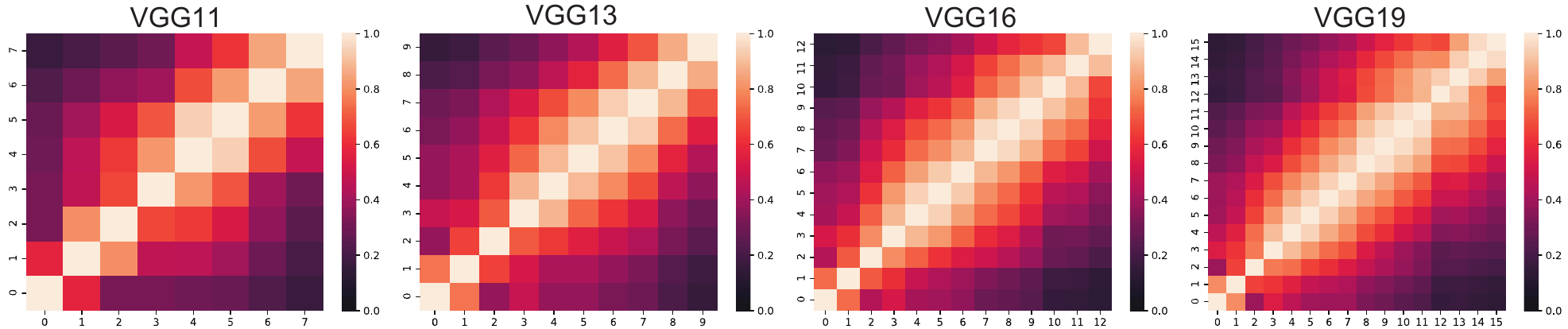}
    \end{subfigure}
    \begin{subfigure}[b]{0.95\textwidth}
            \centering
            \includegraphics[width=\textwidth]{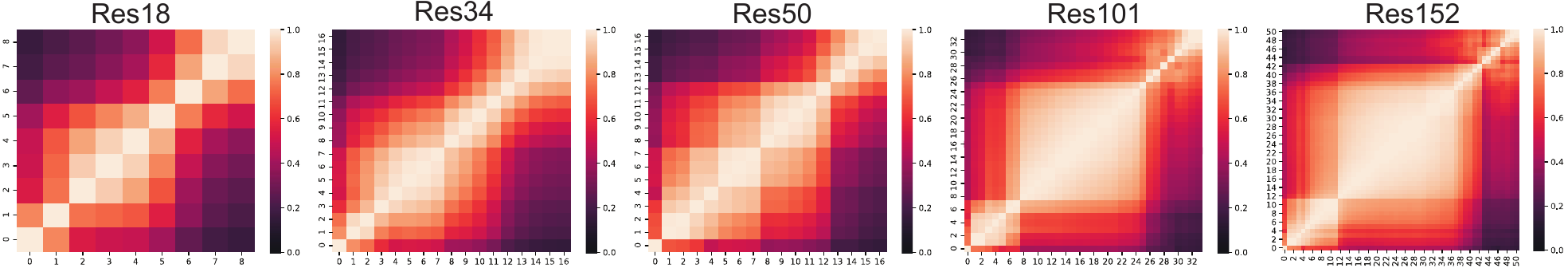}
         
    \end{subfigure}
    \caption{The CKA similarity between the layers of VGGNets and ResNets on CIFAR10.}
    \label{Fig:ckapath_c10}
\end{figure*}

\begin{figure*}[!t]
    \centering
    \begin{subfigure}[b]{0.95\textwidth}
           \centering
           \includegraphics[width=\textwidth]{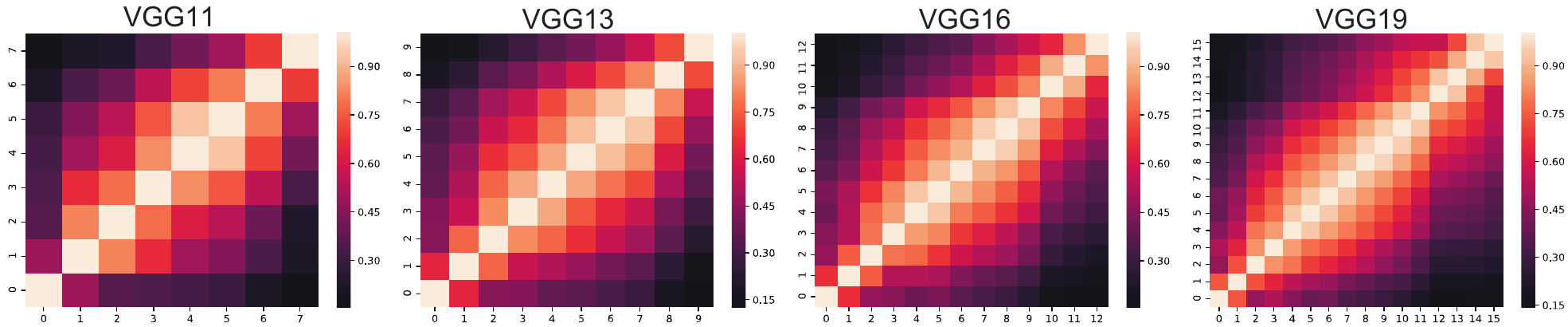}
    \end{subfigure}
    \begin{subfigure}[b]{0.95\textwidth}
            \centering
            \includegraphics[width=\textwidth]{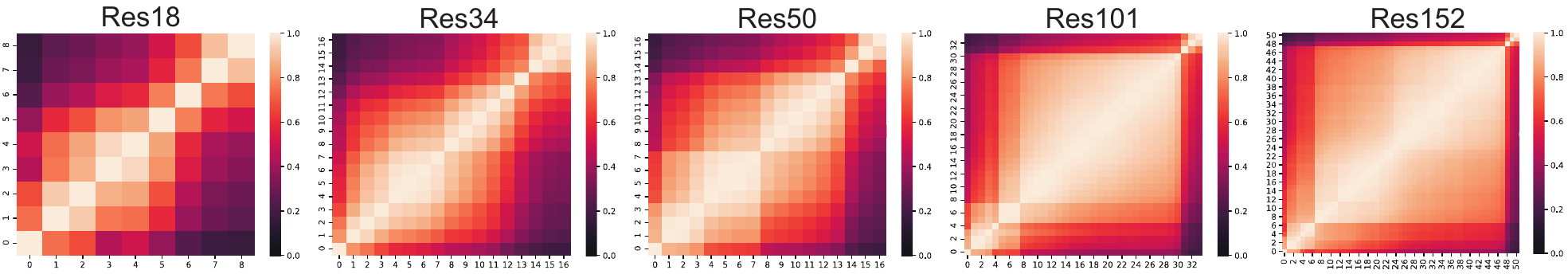}
         
    \end{subfigure}
    \caption{The CKA similarity between the layers of VGGNets and ResNets on CIFAR100.}
    \label{Fig:ckapath_c100}
\end{figure*}

\begin{figure*}[!t]
    \centering
    \begin{subfigure}[b]{0.95\textwidth}
           \centering
           \includegraphics[width=\textwidth]{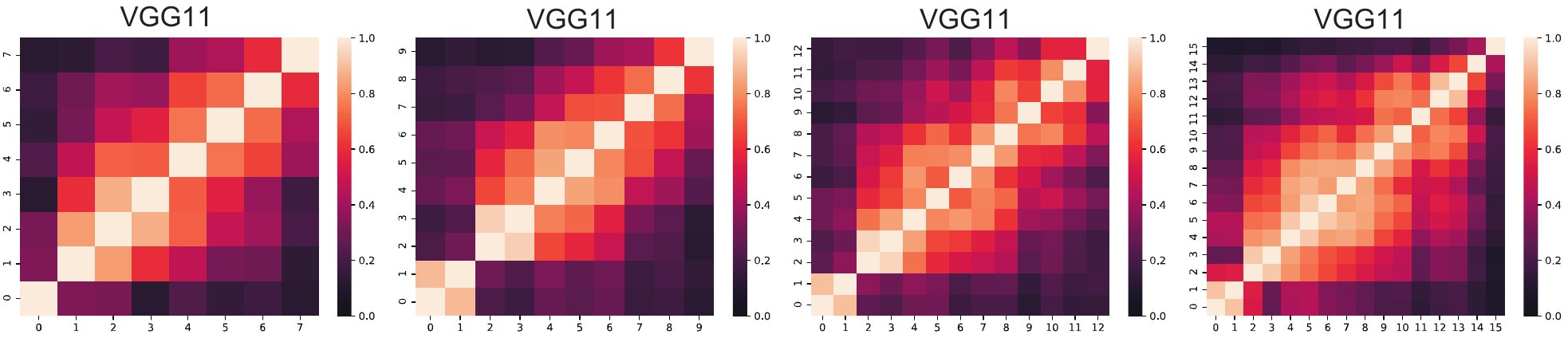}
    \end{subfigure}
    \begin{subfigure}[b]{0.95\textwidth}
            \centering
            \includegraphics[width=\textwidth]{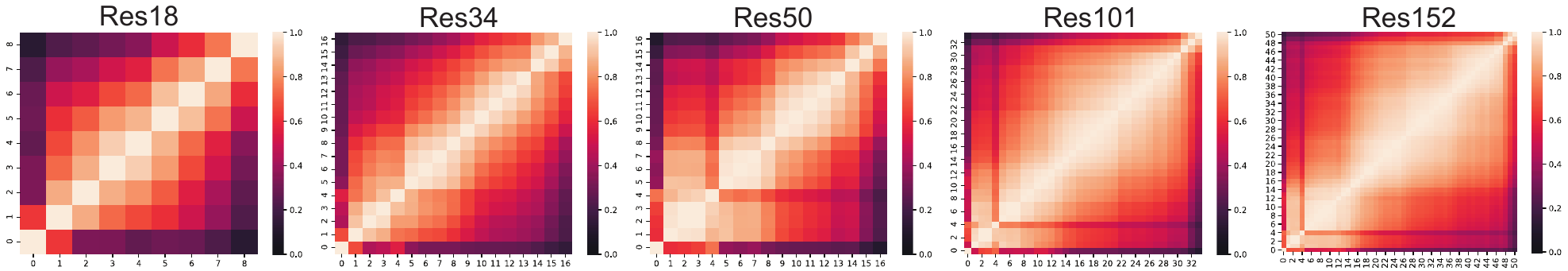}
         
    \end{subfigure}
    \caption{The CKA similarity between the layers of VGGNets and ResNets on ImageNet.}
    \label{Fig:ckapath_imnt}
\end{figure*}

We also conduct the same experiment using CKA on CIFAR10, CIFAR100, and ImageNet. The results are shown in \autoref{Fig:ckapath_c10}, \autoref{Fig:ckapath_c100}, and \autoref{Fig:ckapath_imnt}. CKA also demonstrates grids on the similarity matrix, however, the grids are less clear than GBS, especially on CIFAR100 and ImageNet. For VGGNets on CIFAR10 and CIFAR100, CKA does not show clear grids corresponding to the blocks in VGGNets, while GBS does.

\section{Extensibility of GBS}
In our work, we adopt LSim to measure the similarity of graphs that correspond to hidden representations. There are other graph similarity measurements in addition to LSim, e.g., degree correlation.

The degree of a node is defined as the number of edges connected to it. If we arrange the degrees of $\mathcal{V} \in\mathcal{G}$ into a sequence $\textbf{k}$, then $\textbf{k}$ is called the degree sequence of $\mathcal{G}$. Assume the degree sequence of layer $i$ is $\textbf{k}_i$, which is a $1\times N$ vector, then we can use correlations like Pearson to measure the similarity. The degree correlation of layer $i$ and $j$ is 
\begin{equation}
    \rho_{i,j} = \frac{\mathbb{E}[(\textbf{k}_i-\mu_i)(\textbf{k}_j-\mu_j)]}{\sigma_i \sigma_j},
\end{equation}
where $\mu_i$ is the mean value of $\textbf{k}_i$ and $\sigma_i$ is the variance of $\textbf{k}_i$.

\begin{figure}[!t]
    \centering
    \begin{subfigure}[b]{0.49\textwidth}
           \centering
           \includegraphics[width=0.9\textwidth]{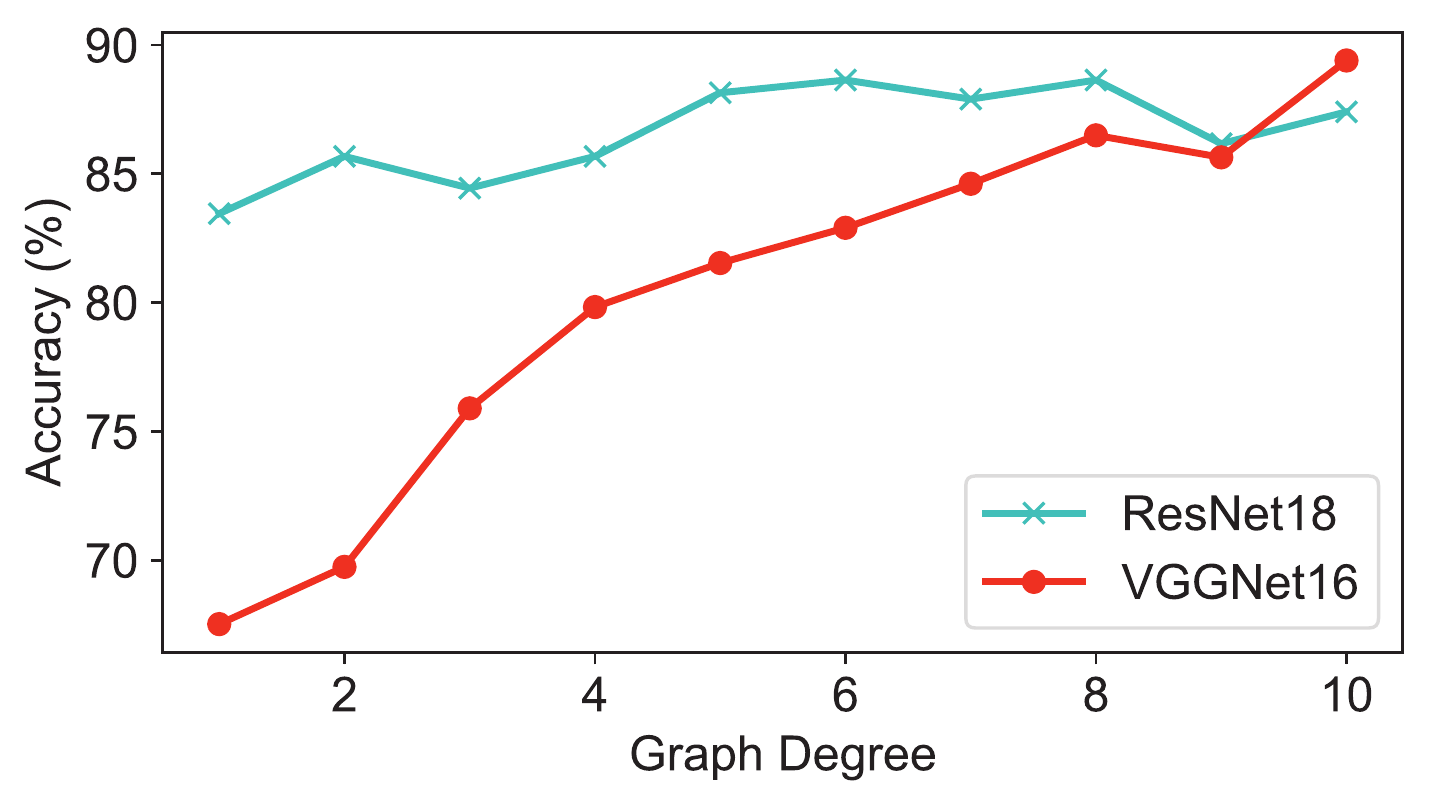}
           \label{Fig:pear_sanitya}
           \caption{}
    \end{subfigure}
    \begin{subfigure}[b]{0.49\textwidth}
            \centering
            \includegraphics[width=0.9\textwidth]{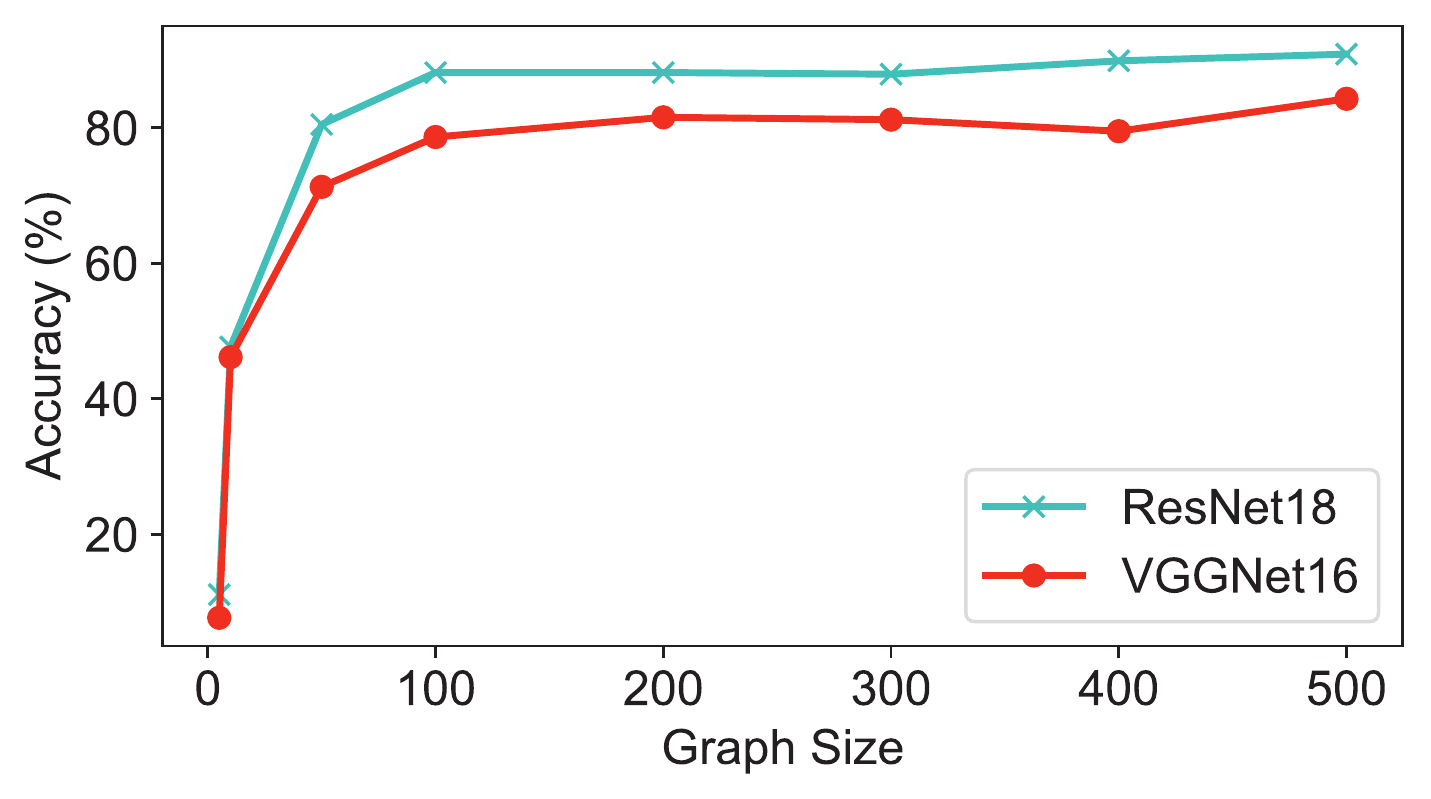}
            \label{Fig:pear_sanityb}
            \caption{}
    \end{subfigure}
    \caption{Sanity check accuracy using Pearson correlation with VGGNet16 and ResNet18 on CIFAR10.}
    \label{Fig:pear_sanity}
\end{figure}

We demonstrate the ablation analysis result of GBS using degree sequence and Pearson similarity, as shown in \autoref{Fig:pear_sanity}. The experiment setting is the same with the cosine similarity. Since the Pearson correlation varies from $-1$ to $+1$, we take the absolute value of the result as the similarity score. Compared with cosine similarity, Pearson correlation shows greater volatility within the same degree range. The accuracy is also lower under the same graph size.

\end{document}